\documentclass[10pt,journal,compsoc]{IEEEtran}

\usepackage{amssymb}
\usepackage[nocompress]{cite}
\usepackage{url}
\usepackage{amsmath}
\usepackage{xcolor}
\usepackage{csquotes}
\MakeOuterQuote{"}
\usepackage{graphicx}
\usepackage{enumitem}
\usepackage{amsfonts}
\usepackage{xpatch}

\usepackage{scrextend}

\usepackage{algorithm}
\usepackage[noend]{algpseudocode}

\usepackage{arydshln}

\usepackage{booktabs}

\usepackage{wrapfig}

\usepackage{mathtools}
\DeclarePairedDelimiter\ceil{\lceil}{\rceil}

\newcounter{mybibstartvalue}
\usepackage{commath}
\usepackage{epstopdf}
\usepackage{graphicx}
\usepackage{blkarray, bigstrut} %
\usepackage{adjustbox}
\usepackage[hidelinks]{hyperref}
\usepackage[acronym]{glossaries}
\usepackage{amsmath, bm}
\begin{document}

	\title{Non-orthogonal Age-Optimal Information Dissemination in Vehicular Networks: A Meta Multi-Objective Reinforcement Learning Approach}

	\author{Ahmed A. Al-Habob,~\IEEEmembership{Member,~IEEE,} 
			Hina Tabassum,~\IEEEmembership{Senior Member,~IEEE,}   and 
		Omer Waqar,~\IEEEmembership{Senior Member,~IEEE}
		\thanks{This work  was
supported by two Discovery Grants funded by the Natural Sciences and
Engineering Research Council of Canada (NSERC).}
\thanks{At the time of this work, A. A. Habob was with the Department of Electrical Engineering and Computer Science at York University, Toronto, Canada. He is currently with the Memorial University of Newfoundland, Canada.}
\thanks{H. Tabassum is with the  Department of Electrical Engineering and Computer Science, York University, Toronto, ON, Canada.}
\thanks{O. Waqar is with the School of Computing, University of the Fraser Valley, BC, Canada and also affiliated as an adjunct faculty member with the Department of Electrical Engineering and Computer Science, York University, Toronto, ON, Canada.  
}
\thanks{E-mails: alhabob@mun.ca, hinat@yorku.ca, Omer.Waqar@ufv.ca.}
}
	
	%\IEEEpeerreviewmaketitle
	\IEEEtitleabstractindextext{
	\begin{abstract}
  This paper considers minimizing the age-of-information (AoI) and transmit power consumption in a vehicular network, where a   roadside unit (RSU) provides timely updates about a set of physical processes to vehicles.  We consider non-orthogonal multi-modal information dissemination, which is based on superposed message transmission from RSU and successive interference cancellation (SIC) at vehicles. The formulated problem is a multi-objective mixed-integer nonlinear programming problem; thus, a Pareto-optimal front is very challenging to obtain. First, we leverage the weighted-sum approach to decompose the multi-objective problem into a set of multiple single-objective sub-problems corresponding to each predefined objective preference weight. Then, we develop a hybrid deep Q-network (DQN)-deep deterministic policy gradient (DDPG) model to solve each optimization sub-problem respective to predefined objective-preference weight. The DQN optimizes the decoding order, while the DDPG solves the continuous power allocation. The model needs to be retrained for each sub-problem. We then present a two-stage meta-multi-objective reinforcement learning solution to estimate the Pareto front with a few fine-tuning update steps without retraining the model for each sub-problem. Simulation results illustrate the efficacy of the proposed solutions compared to the existing benchmarks and that the meta-multi-objective reinforcement learning model estimates a high-quality Pareto frontier with reduced training time.
	\end{abstract}
	
	\begin{IEEEkeywords}
		Age-of-information (AoI),   
		deep reinforcement learning (DRL),  meta deep reinforcement learning (meta-DRL), multi-objective optimization, successive interference cancellation (SIC).
	\end{IEEEkeywords}
}
\maketitle
\IEEEdisplaynontitleabstractindextext
\IEEEdisplaynontitleabstractindextext

% \fi
\IEEEpeerreviewmaketitle

\IEEEraisesectionheading{\section{Introduction}\label{sec:introduction}}
 {	\IEEEPARstart{V}{ehicular} communication networks 
 % are envisioned  to play a critical role in the future transportation systems to make everyday vehicular operation greener, safer,  and more efficient, especially for intelligent driving assistance and travel experience enhancement. 
enable a wide range of   applications  which  require real-time updates, such as  highly prioritized road safety applications (such as collision warning and adaptive cruise control) and the  infotainment services such as news, media and social entertainments, which  require real-time updates  \cite{8466351}. 
With the increasing diversity of vehicular applications that require real-time information updates, such as   blind spot/lane change   and forward collision warnings, communications in vehicular networks become time-critical, and thus,  fresh status  updates are of   high importance \cite{9730060}. 

Although the conventional communication latency and throughput are effective metrics to evaluate the performance  of the vehicular communication networks, these  
metrics do not capture the information freshness which is critical to obtain the real-time knowledge about the location, orientation, and speed of the  vehicles. To this end, the age-of-information (AoI) is  a useful metric to quantify the  freshness of the information while taking into account  the transmission latency, update generation time,  and inter-update time interval. Specifically, AoI is defined as the elapsed time between the  received information at the destination and the time when it was generated at the source \cite{6195689}. It should be noted that the inter-update time ---which is  a scheduling parameter--- is a crucial parameter in the AoI \cite{6195689}, and thus  optimizing AoI is totally different from optimizing other metrics such as the throughput and latency.

Along another note, the existing state-of-the-art considers minimizing the AoI in uni-modal information dissemination scenario, where the destination/vehicle receives updates about a single application or physical process. However,  a more practical scenario that should be addressed is the multi-modal information dissemination,  in which each vehicle is interested in maintaining fresh status updates for  one or more   applications (physical processes). The straightforward strategy  is to broadcast updates about  all physical processes to all vehicles at each time slot. However, such a  strategy consumes the transmitter's power. Consequently, the
trade-off between  maintaining information freshness and reserving the power consumption should be handled to satisfy the
decision-maker's preference.
   Such a trade-off mandates a proper multi-objective optimization framework that  handles  the decision-maker's preference by 
considering efficient messages' encoding scheme to unicast/multicast updates and allocating the  power    in vehicular networks.}

Traditionally, optimization approaches tackle the resource
allocation  in vehicular networks \cite{7289470}. However, the communication channels in vehicular networks are rapidly varying. Furthermore,  optimizing time-dependent metrics such as AoI involves a sequential
decision-making  over time.  Thus, iterative and computationally complex  optimization approaches are not well-suited. To this end, deep reinforcement learning (DRL) has been considered as a promising solution to  learn
a better policy on decision-making problems that are evolving over time\cite{8954939,9195789}. 

DRL paradigm has witnessed dramatic evolution to handle   complex scenarios such as   discrete-continuous
hybrid action space and different training and testing environments \cite{9495238}. 
 A variant of the DRL  is the multi-objective reinforcement learning (MORL) \cite{6918520}, in which   the agent aims at optimizing
 multiple conflicting objectives. MORL inherits the well-known challenges of multi-objective optimization\footnote{The conventional strategy to handle multi-objective optimization is to convert
 the problem into a single objective optimization problem, which can be implemented using   
 $ \epsilon $-constraint approach \cite{9131855},      Tchebycheff approach   \cite{9485101}, or weighted sum approach \cite{kim2006adaptive,marler2004survey,augusto2012new}.}, including    the trade-off between   objectives with different units, ranges, and order of magnitude \cite{marler2004survey}.

 To address the aforementioned issues, a DRL agent can be trained and tested to find  the preferred Pareto optimal solution based on the decision-maker's preference.  The drawback is that the DRL agent needs to be retrained  if the objectives' preference is modified or the vehicular communication
environment is changed. 
  %In this case, a DRL agent    should be trained for  a specific   objectives' preference weight, which is ineffective as it leads to a large number of DRL agents. 
  Recently, meta-DRL concept 
   has been   introduced  to enable the agent to quickly adapt to new environments by learning the agent a meta-policy that solves multiple tasks from a given distribution \cite{finn2017model}. Meta-learning (or the learning to
   learn) can leverage
   knowledge from previous experiences to rapidly adapt to new/unseen tasks with few
   training (fine-tuning) steps \cite{finn2017model,9714721}.   
    In this context, meta-DRL can be considered a posteriori paradigm to address the  objectives' preference issue, in which the  meta-DRL is trained to find the Pareto optimal solutions of a set of objectives' preferences which can be adapted or fine-tuned to the  desirable solution \cite{chen2019meta,9714721}.

\subsection{Related Works}  
Optimizing AoI in vehicular networks has been investigated in different scenarios    \cite{8424604,8937801,8954939,9149054,9730060}.
In \cite{8424604}, a   greedy algorithm was developed to minimize 
the expected sum AoI in a vehicular beacon broadcasting system and mitigate beacons' signals collision. 
 In \cite{8937801}, a Lyapunov optimization solution was developed to minimize the transmit power in a vehicle-to-vehicle (V2V) network to facilitate ultra-reliable low-latency V2V communications subject to probabilistic AoI constraints. A proactive DRL technique  was proposed in
\cite{8954939} to provide an 
AoI-aware radio resource
management in vehicular networks with Manhattan
grid road topology. In \cite{9149054},
the impact of inter-update generation, selection of fog/cloud servers, and   
processing delay  on the AoI was studied
in a  vehicular   shuttle system.   A deep Q-learning algorithm was developed to
optimize the  vehicles' routing to improve the
average AoI.
In \cite{9730060}, the social relations
among users in the vehicle network were   considered   in  an AoI-centric information
dissemination  model.  The authors considered   
joint optimization of the information update rate  
and the transmit probabilities   to minimize 
AoI.

Different unicast–multicast scenarios were considered in literature to transmit multiple messages to a group of users \cite{8048611,8417939,8828094,8412246,9379910}. 
In \cite{8048611}, a non-orthogonal multiple access (NOMA) unicast–multicast scenario was designed, in which a set  of
unicast users (each requires unique message) and a  set
of multicast users (those who require an identical message) shares the
same time/space/frequency resource. In this scenario, the messages of unicast users were encoded   according to their channels' quality and the  multicast message  was superposed as the last encoded message.
An   integrated multicast-unicast scenario was   proposed  in \cite{8417939,8828094}, such that each   user receives a private message and a common message was broadcasted to all users. Sum-rate maximization was considered in \cite{8417939}, while the authors of 
\cite{8828094} considered maximizing the energy efficiency in a   simultaneous  wireless information and power transfer   protocol. 
%%%%%%%%%%%%%%%%%%%%%%%%%%%%
In \cite{8412246}, a joint unicast and multi-group multicast transmission has been considered,  in  which a user 
is
either a unicast  or belongs to a group of   multicast  users.   In \cite{9379910}, a 
  fixed interval transmission
strategy was considered to minimize  the average AoI in a unilateral multicast network. The considered network consists of a base-station (BS)  broadcasting time-sensitive updates  to a set of users, in which the  BS decides to
broadcast an update  packet at time. {\color{black}In  \cite{9899364},  a digital twin-driven vehicular task offloading was studied to provide augmented computing
	capacities for internet of vehicles (IoV) scenario by  considering mobile edge
	computing (MEC) and intelligent reflective surface
	(IRS).   To reduce the overall delay and energy
	consumption,   a two-stage optimization approach was developed
	for Jointly Optimizing Task Offloading and IRS Configuration. Another IoV scenario was studied in \cite{yuan2023low}, in which  an asynchronous federated broad learning (FBL) framework   integrates broad learning (BL) into federated learning (FL).  }

Studying AoI in unicasts/multicasts scenarios has been  considered in literature \cite{9037357,8422521,9465806,10110908,al2022dynamic}.
In \cite{9037357},  the average and peak AoI of multicast
transmission with deadlines has been considered. An access point transmits timestamped status
updates to multiple devices and the status
update is terminated   if either a subset of devices devices successfully receive the status update or the deadline expires. In \cite{8422521}, a  multicast scheduling strategy has been proposed to improve the energy   efficiency and AoI. The   server transmits information to multiple users by  queuing and bundling the requests from different users   and serving   users requesting the same contents. In \cite{9465806},  the AoI in multicast networks with retransmissions has been studied, in which   the updates are encoded as a short blocklength packets and is broadcasted to multiple destinations via independent and identically distributed error-prone channels. The stopping threshold, the average AoI and EE expressions for stopping   and wait-for-all schemes are derived as a function of the packet length.  In \cite{10110908}, an architecture of AoI in multicast/unicast/device-to-device (D2D) transmission with a   cell-free
massive multiple-input multiple-output (MIMO) was studeied.  An   age-optimum unicast-multicast scheduling of multiple update messages to vehicles   has   been considered in \cite{al2022dynamic}.  In this framework, at most     an update about one physical process is scheduled to a vehicle at each time slot. Ant colony optimization algorithm and deep Q-learning model were developed to solve a weighted sum   optimization problem of the AoI and power consumption.

\begin{table*}[h!]
	\begin{center}\color{black}
		\caption{Existing related work.}
		\label{Related_Table}
		\begin{adjustbox}{width=.99\textwidth}
			\begin{tabular}{|c|c|c|c|c|c|c|c|c|}
				\hline
				{Ref.} &  {Year} &  {Vehicular network}&  {AoI} & {NOMA  multi-modal dissemination} & {DRL-based solution}   & {Hybrid DQN-DDPG}  & meta-DRL  & { Pareto front using meta-MORL} \\ \hline 
				\cite{8424604}	&\multicolumn{1}{c|}{$ 2018 $ }  & \checkmark & \multicolumn{1}{c|}{\checkmark}     &   \multicolumn{1}{c|}{{X}}  &  \multicolumn{1}{c|}{X} &  \multicolumn{1}{c|}{X} &  \multicolumn{1}{c|}{X} &  \multicolumn{1}{c|}{X} \\ \hline
				\cite{8937801}	&\multicolumn{1}{c|}{$ 2020 $ }  & \checkmark & \multicolumn{1}{c|}{\checkmark}     &   \multicolumn{1}{c|}{{X}}  &  \multicolumn{1}{c|}{X} &  \multicolumn{1}{c|}{X} &  \multicolumn{1}{c|}{X} &  \multicolumn{1}{c|}{X} \\ \hline
				\cite{8954939}	&\multicolumn{1}{c|}{$ 2020 $ }  & \checkmark & \multicolumn{1}{c|}{\checkmark}     &   \multicolumn{1}{c|}{{X}}  &  \multicolumn{1}{c|}{\checkmark} &  \multicolumn{1}{c|}{X} &  \multicolumn{1}{c|}{X} &  \multicolumn{1}{c|}{X} \\ \hline
				\cite{9149054}	&\multicolumn{1}{c|}{$ 2020 $ }  & \checkmark & \multicolumn{1}{c|}{\checkmark}     &   \multicolumn{1}{c|}{{X}}  &  \multicolumn{1}{c|}{\checkmark} &  \multicolumn{1}{c|}{X} &  \multicolumn{1}{c|}{X} &  \multicolumn{1}{c|}{X} \\ \hline
				\cite{9730060}	&\multicolumn{1}{c|}{$ 2022 $ }  & \checkmark & \multicolumn{1}{c|}{\checkmark}     &   \multicolumn{1}{c|}{{X}}  &  \multicolumn{1}{c|}{{X}} &  \multicolumn{1}{c|}{X} &  \multicolumn{1}{c|}{X} &  \multicolumn{1}{c|}{X} \\ \hline
				\cite{9037357}	&\multicolumn{1}{c|}{$ 2020 $ }  & {X} & \multicolumn{1}{c|}{\checkmark}     &   \multicolumn{1}{c|}{{X}}  &  \multicolumn{1}{c|}{{X}} &  \multicolumn{1}{c|}{X} &  \multicolumn{1}{c|}{X} &  \multicolumn{1}{c|}{X} \\ \hline
				\cite{8422521}	&\multicolumn{1}{c|}{$ 2018 $ }  & {X} & \multicolumn{1}{c|}{\checkmark}     &   \multicolumn{1}{c|}{{X}}  &  \multicolumn{1}{c|}{{X}} &  \multicolumn{1}{c|}{X} &  \multicolumn{1}{c|}{X} &  \multicolumn{1}{c|}{X} \\ \hline
				\cite{9465806}	&\multicolumn{1}{c|}{$ 2021 $ }  & {X} & \multicolumn{1}{c|}{\checkmark}     &   \multicolumn{1}{c|}{{X}}  &  \multicolumn{1}{c|}{{X}} &  \multicolumn{1}{c|}{X} &  \multicolumn{1}{c|}{X} &  \multicolumn{1}{c|}{X} \\ \hline
				\cite{10110908}	&\multicolumn{1}{c|}{$ 2023 $ }  & {X} & \multicolumn{1}{c|}{\checkmark}     &   \multicolumn{1}{c|}{{X}}  &  \multicolumn{1}{c|}{{X}} &  \multicolumn{1}{c|}{X} &  \multicolumn{1}{c|}{X} &  \multicolumn{1}{c|}{X} \\ \hline
				\cite{9495238}	&\multicolumn{1}{c|}{$ 2021 $ }  & {\checkmark} & \multicolumn{1}{c|}{{X}}     &   \multicolumn{1}{c|}{{X}}  &  \multicolumn{1}{c|}{{\checkmark}} &  \multicolumn{1}{c|}{\checkmark} &  \multicolumn{1}{c|}{\checkmark} &  \multicolumn{1}{c|}{X} \\ \hline
	        	\cite{al2022dynamic}	&\multicolumn{1}{c|}{$ 2022 $ }  & {\checkmark} & \multicolumn{1}{c|}{\checkmark}     &   \multicolumn{1}{c|}{{X}}  &  \multicolumn{1}{c|}{{\checkmark}} &  \multicolumn{1}{c|}{X} &  \multicolumn{1}{c|}{X} &  \multicolumn{1}{c|}{X} \\ \hline
	        	Ours	&\multicolumn{1}{c|}{ --}  & {\checkmark} & \multicolumn{1}{c|}{\checkmark}     &   \multicolumn{1}{c|}{{\checkmark}}  &  \multicolumn{1}{c|}{{\checkmark}} &  \multicolumn{1}{c|}{\checkmark} &  \multicolumn{1}{c|}{\checkmark} &  \multicolumn{1}{c|}{\checkmark} \\ \hline
			\end{tabular}
		\end{adjustbox}
	\end{center}
\end{table*}

\subsection{Motivation and Contributions}
{None of the research works to date considered the \textit{non-orthogonal multi-modal information dissemination} in a vehicular network with \textit{multiple conflicting objectives such as AoI minimization and power consumption minimization simultaneously}. We cast this problem as a multi-objective optimization problem (MOOP) and  demonstrate the efficacy and the  generalization capability of the \textit{meta multi-objective RL}  for estimating the \textbf{Pareto front} of the MOOP. Estimating the Pareto front in MOOP is essential, as it enables the decision-maker to optimize the conflicting objectives without the need to pre-determine the preference weight corresponding to each Pareto point. It is shown that the proposed model can estimate the entire Pareto front using few fine-tuning update steps, {without the need to retrain a new DRL model for each point in the Pareto front}. {\color{black} To summarize the existing work and highlight   our key contributions, Table \ref{Related_Table} summarizes the related work in terms of considering key techniques such as minimizing AoI, meta-DRL, meta-MORL, etc. }
	The main contributions of this paper can then be summarized as follows:

{	\color{black}
	\begin{itemize}
		\item This paper proposes a \textit{non-orthogonal} multi-modal information dissemination framework\footnote{The non-orthogonal multi-modal information dissemination  is based on superposed message transmission from RSU and successive interference cancellation (SIC)-enabled decoding at vehicles. }  in which each vehicle can receive updates about \textit{one or more physical processes at a time}. A roadside unit (RSU)  schedules updates on multiple processes such that the \textit{average AoI  and the RSU's power consumption} can be minimized at the same time. The two objectives are coupled in a conflicting manner due to the transmit power allocations.
		
		\item We develop a meta multi-objective reinforcement learning (meta-MORL) framework  to minimize both the AoI at the vehicles and the RSU’s power expenditure while optimizing the messages' decoding order and   their corresponding power allocations. In this context, we  
  \begin{itemize}
      \item first design a   hybrid  DRL  model, namely,   hybrid deep  Q-network
	(DQN)-deep deterministic policy gradient (DDPG)  model to obtain the Pareto front of the considered multi-objective problem.  The DQN   solves the   messages' decoding order
		and  the DDPG handles the continuous power allocation decision. The model needs to be retrained for each point of the Pareto front.
  \item then, we develop a  two-stage meta-MORL  solution to deal with the multiple sub-problems determined by the preference weight  of the objectives and to estimate the Pareto front \textit{without retraining}. The first stage  trains a policy with a good generalization of the preference weight  of the objectives. The fine-tuning stage is then applied to quickly adapt the trained policy for  an unseen preference weight  of the objectives.
  \end{itemize}
		\item Extensive simulations are provided to evaluate the   performance of the  proposed algorithm and the generalization capability of the proposed  meta-MORL solution. The results demonstrate that the proposed algorithms can efficiently optimize  the messages' decoding order and  
		power allocation issues, and the meta-multi-objective RL adapts quickly to
		the   problem instances with  unseen/new objective-preference weight.      
\end{itemize}}
}

The remainder of this paper is organized as follows.
Section~\ref{Sys} presents the  system model and the performance metrics. The problem is formulated in Section \ref{Prob}.    The proposed hybrid DQN-DDPG  DRL model is introduced in  Section \ref{DRLApproach} and the multi-objective meta-DRL solution is      introduced in  Section \ref{METADRL}.  Section~\ref{SimSec}  illustrates
simulation results  and Section~\ref{Con} concludes the paper.

\begin{table*}[h!]
	\begin{center}
		\caption{Main notations used in the paper.}
		\label{my-label}
		\begin{adjustbox}{width=1\textwidth}
			\begin{tabular}{|c|c||c|c|}
				\hline
				{Notation} &  {Description} & {Notation} &  {Description}\\ \hline 
				$ V $	&\multicolumn{1}{l||}{Number of vehicles in the vehicles set $ \mathcal{V}  $ }  & $ N $ & \multicolumn{1}{l|}{Number of antenna elements at the RSU}           \\ \hline
				$F $	&\multicolumn{1}{l||}{Number of   physical processes in the processes set $ \mathcal{F}  $}      & $ L $  &\multicolumn{1}{l|}{Payload size of  an update     (in bits)}         \\ \hline
				$  \mathcal{R}_i   $	&\multicolumn{1}{l||}{Processes  of interest to vehicle $ v_i$ ($  \mathcal{R}_i \subseteq \mathcal{F}  $) }         & $\bm{R}=[r_{i,l}]_{\footnotesize{ V\times F}} $  &\multicolumn{1}{l|}{$ 	r_{i,l} =1 $ if vehicle $v_i$ is interested  in process $f_l$, $ r_{i,l} =0 $ otherwise}         \\ \hline
				$ T/\delta $		&\multicolumn{1}{l||}{Number of time slots/Duration of each slot }            &$ \psi_0=\{x_0,y_0\} $  & \multicolumn{1}{l|}{Coordinates of the RSU}             \\ \hline
				$ \psi^{(t)}_i=\{x^{(t)}_i,y^{(t)}_i\} $		
				&\multicolumn{1}{l||}{Coordinates of vehicle $ i $ at time slot $ t $}               &$ \phi_i^{(t)} $  &\multicolumn{1}{l|}{Angle of vehicle $ i $ relative to the RSU at time slot $ t $}   \\ \hline
				$ \ell_i^{(t)} $		&\multicolumn{1}{l||}{Distance between the   vehicle $ i $ and the RSU}            & $ 	\mbox{\textbf{h}}_i^{(t)} $ &\multicolumn{1}{l|}{Communication channel between the RSU and vehicle $i$ at time slot $t$}              \\ \hline
				$ f_c/c_0/c_i  $		
				&\multicolumn{1}{l||}{Carrier frequency/Speed of light/Speed of vehicle $ i $ }               &$ \varrho_i^{(t)} $  &\multicolumn{1}{l|}{Doppler shift due to the movement of   vehicle $ i $}                 \\ \hline
				$\mbox{\textbf{a}}(\phi_i^{(t)})/\phi_i^{(t)}$	 &\multicolumn{1}{l||}{Transmit steering vector/Azimuth angle between  RSU and  $v_i$ at  slot $t$}             &$ \bm{\pi}^{(t)} $  &\multicolumn{1}{l|}{Decoding order decision of the messages at time slot $t$}     \\ \hline
				$\mbox{\textbf{p}}^{(t)}$	& \multicolumn{1}{l||}{Power allocation decision at time slot $t$}        &$ z_i^{(t)}/n_i $  &\multicolumn{1}{l|}{Received signal/Additive white Gaussian noise  at vehicle  $i$}          \\ \hline
				$\mbox{\textbf{{w}}}$	&  \multicolumn{1}{l||}{Beamforming vector at time slot $t$}        &$\chi^{(t)}_{i}/\tilde{\xi}$  &\multicolumn{1}{l|}{Large-scale channel attenuation of vehicle $i$/Normalization factor}           \\ \hline
				$\gamma_{i,\pi_{l'}^{(t)}}^{(t)}(\bm{\pi}^{(t)},\mbox{\textbf{p}}^{(t)})$	&  \multicolumn{1}{l||}{SINR experienced at vehicle $i$ 
					to decode the $ \pi_{l'}^{(t)} $-th message}        & $ \varepsilon^{\mbox{\scriptsize max}}_i/	\varepsilon_i(\gamma_i^{(t)})  $ &\multicolumn{1}{l|}{Maximum allowed error probability/Decoding error probability } 
				\\ \hline
				$\Phi\left(\cdot\right)/\omega$	&  \multicolumn{1}{l||}{Channel
					dispersion/Channel bandwidth}        &$ \delta_1/\delta_2 $  &\multicolumn{1}{l|}{Vehicles' parameters acquisition time/Information transmission time} 
				\\ \hline
				$	\Delta_{i\pi_{l'}}^{(t)}/\bar{\Delta}_{{i,l}}$	&  \multicolumn{1}{l||}{Instantaneous AoI/Time-average  AoI of $f_l$ at vehicle  $ i $}        &$ \bar{\Delta}^{\mbox{\scriptsize max}}/\bar{\Delta}^{\mbox{\scriptsize min}}$  &\multicolumn{1}{l|}{Maximum/Minimum value of the total time-average  AoI}          \\ \hline
				$\bar{{p}}^{(t)}$	&  \multicolumn{1}{l||}{Time-average   power consumption at RSU}        &${O}(\bm{\pi}^{(t)}\!,\mbox{\textbf{p}}^{(t)})/\zeta $  &\multicolumn{1}{l|}{Objective function/Relative objective's preference weight}          \\ \hline
				$ \mathcal{S}/\mathcal{A}/\rho^{(t)} $	&  \multicolumn{1}{l||}{State space/Action space/Immediate reward}        &$ \theta^Q/\theta^{Q_c}/\theta^\mu $  &\multicolumn{1}{l|}{DQL network weights/Critic   network weights/Actor network weights}         
				\\ \hline
			\end{tabular}
		\end{adjustbox}
	\end{center}
\end{table*}

 \section{System Model and  Performance Metrics}\label{Sys}
 This section  introduces the considered system model,
 	communication model, and performance metrics. { The notations used throughout this
 	paper are listed in Table \ref{my-label}.}

 \subsection{Network Model}
The considered system  consists of a  set $\mathcal{V} = \{v_i\}_{i=1}^{V}$ of  $ V $ vehicles supported by an   RSU  that   disseminates  timely
status updates  to  the vehicles.  The RSU is equipped with
a   uniform linear array  of $ N $ antennas. A multi-modal data dissemination scenario is considered, in which the RSU  is capable of providing timely
status updates  about a set $\mathcal{F} = \{f_l\}_{l=1}^{F}$ of $ F $ physical processes. The payload size of  an update     is   $L$ bits.{ \color{black} Each vehicle is interested in   maintaining   
freshness of its information
status about a subset of    processes $ \mathcal{R}_i \subseteq \mathcal{F} $. To represent the information demands of the   vehicles, we define $ \bm{R}=[r_{i,l}]_{\footnotesize{ V\times F}} $   such that 
\begin{equation}
	r_{i,l} = \begin{cases} 1, & \mbox{if vehicle } i  ~\mbox{is interested  in process}~ l,  \\ 
		0, & \mbox{otherwise}. \end{cases}
\end{equation}}
The time is divided into $ T $ time slots each of duration $ \delta $. Let $ \psi_0=\{x_0,y_0\} $  be the coordinates of the RSU and $ \psi^{(t)}_i=\{x^{(t)}_i,y^{(t)}_i\} $ be the coordinates of vehicle $ i $ at time slot $ t $. The angle of vehicle $ i $ relative to the RSU at time slot $ t $ can be expressed as follows:
\begin{equation}
	\phi_i^{(t)}=\arccos \frac{x^{(t)}_i-x_0}{\ell_i^{(t)}}, 
\end{equation}
where $ \ell_i^{(t)} = \lVert\psi^{(t)}_i-\psi_0\rVert $ is the distance between the   vehicle $ i $ and the RSU.
{
The communication channel between the RSU and vehicle $i$ at time slot $t$ is modeled as follows:
\begin{equation}
	\mbox{\textbf{h}}_i^{(t)} = \sqrt{\frac{c_0}{4\pi f_c\ell_i^{(t)^2}}}\mbox{\textbf{a}}^H(\phi_i^{(t)}) e^{j2\pi \varrho_i^{(t)}},
\end{equation} 
where $ f_c $ is the carrier frequency,  $ c_0 $ is the speed of light,  and  $ \varrho_i^{(t)} $ is the
Doppler shift due to the movement of   vehicle $ i $ expressed as
$
\varrho_i^{(t)}=\frac{c_i f_c\cos \phi_i^{(t)} }{c_0}, 
$
where $ c_i $ is the speed of vehicle $ i $ \cite{9557830}. Assuming a uniform linear antenna array at the RSU, the transmit array steering vector  $ \mbox{\textbf{a}}(\phi_i^{(t)} ) \in  \mathbb{C}^{N\times 1} $ (with  $ \phi_i^{(t)} $ as the   azimuth angle between the RSU and vehicle $i$ at time slot $t$) can be expressed as  follows:
\begin{equation}
	\mbox{\textbf{a}}(\phi_i^{(t)})\! = \! [1,e^{j\pi \sin \phi_i^{(t)}}\!, e^{j 2\pi\sin \phi_i^{(t)} }\!,  \cdots\!, e^{j (N-1)\pi\sin \phi_i^{(t)} }],
\end{equation}
where $ j=\sqrt{-1} $ and the antenna spacing is $ \lambda/2 $ with $ \lambda $ as the carrier wavelength.}

\subsection{Received Signal and SINR Model}

 Let $[f^{(t)}_1, f^{(t)}_2, \cdots, f^{(t)}_F]$ be the raw messages of the physical processes $\mathcal{F}$ at time slot $t$, {\color{black} the RSU sends a superposed message $ \sum_{l=1}^{F}\sqrt{p^{(t)}_{\pi_l^{(t)}}}f^{(t)}_{\pi_l^{(t)}} $ according to a decoding order decision $ \bm{\pi}^{(t)} = [\pi^{(t)}_1, \pi^{(t)}_2, \cdots, \pi^{(t)}_{F}] $ and power allocation decision $ \mbox{\textbf{p}}^{(t)} =[p^{(t)}_1, \cdots, p^{(t)}_F] $,  where $\pi^{(t)}_l$ is the $l$-th elements of the decoding order decision $\bm{\pi}^{(t)}$. The RSU broadcasts the superposed message   to the
vehicles. Therefore, the received signal at vehicle  $i$ can be modeled as follows:

\begin{equation}
	z_i^{(t)} = \mbox{\textbf{{w}}}^{(t)} \left(\sum_{l=1}^{F}\sqrt{p^{(t)}_{\pi_l^{(t)}}}f^{(t)}_{\pi_l^{(t)}}\right)\mbox{\textbf{h}}_i^{(t)^H} +n_i,
\end{equation}}
where $n_i \sim \mathcal{CN}(0, \sigma^2)$ is the additive white Gaussian noise (AWGN), $H$ denotes the Hermitian transpose, and $ \mbox{\textbf{{w}}}^{(t)} $ is the beamforming vector.  The maximum ratio transmission (MRT)   
beamforming scheme is considered \cite{8317005}, in which the asymptotically optimal beamformer vector for  a set of vehicles $\mathcal{V}$    is a linear
combination of channels of these vehicles \cite{7417526,8412246}. Consequently, the MRT beamforming vector   is expressed as  follows:
\begin{equation}
	{\mbox{\textbf{{w}}}}^{(t)}=\sum_{i=1}^{V}	\frac{\mbox{\textbf{h}}_i^{(t)}}{\sqrt{N\chi^{(t)}_{i}\tilde{\xi}}},
\end{equation}
where $ \chi^{(t)}_{i}= {\frac{c_0}{4\pi f_c\ell_i^{(t)^2}}} e^{-j2\pi \varrho_i^{(t)}} $ is the   large-scale channel attenuation of vehicle $i$ and  $ \tilde{\xi} $ is a normalization factor \cite{8412246,8125756}.  
{For a given decoding order decision  $\bm{\pi}^{(t)}$, the vehicles decode the messages such that the
message corresponding to the $ \pi_{l'}^{(t)} $-th physical process  is decoded  before the message of the  $ \pi_{m'}^{(t)} $-th process, $\forall~ {l'}\leq {m'}$. 
Consequently,  the signal-to-interference plus-noise ratio (SINR) experienced at vehicle $i$ 
 to decode the $ \pi_{l'}^{(t)} $-th message    is
given by

\begin{equation} \label{SNR_NOMA_1}
	\gamma_{i,\pi_{l'}^{(t)}}^{(t)}(\bm{\pi}^{(t)},\mbox{\textbf{p}}^{(t)})=\frac{p^{(t)}_{\pi_{l'}^{(t)}}|\mbox{\textbf{h}}_{i}^{(t)^H}{\mbox{\textbf{{w}}}}^{(t)}|^2}{\sum\limits_{\substack{{m'}={l'}+1}}^{F}p^{(t)}_{\pi_{m'}^{(t)}}|\mbox{\textbf{h}}_{i}^{(t)^H}{\mbox{\textbf{{w}}}}^{(t)}|^2+\sigma^2},  \forall 1 \leq {l'}\leq F.
\end{equation}

\subsection{Decoding Error Probability}
{Note that SIC is performed at the vehicles to obtain the required updates, such that  to    successfully estimate the $ \pi_{l'}^{(t)} $-th message     at vehicle $i$, it has to first perform SIC to estimate and remove all the messages $ \pi_{m}^{(t)} $ $ \forall 1 \leq m \leq  l'-1$. Once these messages are removed, vehicle $i$  can perform direct decoding  by treating the data of messages   $ \pi_{l'+1}, \pi_{l'+2}, \cdots F$   as interference according to (7). Consequently, the $ \pi_{l'}^{(t)} $-th message can be  estimated at   vehicle $i$ if

	\begin{equation}\label{Condd}
		\varepsilon_i(\gamma_{i,\pi_{m}^{(t)}}^{(t)}(\bm{\pi}^{(t)},\mbox{\textbf{p}}^{(t)}))\!\leq\varepsilon^{\mbox{\scriptsize max}}_i, ~ \forall ~ 1 \leq m \leq {l'},
	\end{equation}
	where $ \varepsilon^{\mbox{\scriptsize max}}_i $ is the maximum allowed error probability at vehicle $i$ and 	{\color{black} the decoding error probability   can be expressed as follows \cite{9369424}:
	\begin{equation}\label{error}
		\begin{split}
		&	\!\!\!\varepsilon_i(\gamma_{i,\pi_{m}^{(t)}}^{(t)}(\bm{\pi}^{(t)},\mbox{\textbf{p}}^{(t)}))=\\ 
		&\!\!\! \Phi\left(\sqrt{\frac{\delta_2 \omega}{\Gamma^{(t)}_{i,\pi_{m}^{(t)}}}}\left[\ln\left(1+\gamma_{i,\pi_{m}^{(t)}}^{(t)}(\bm{\pi}^{(t)},\mbox{\textbf{p}}^{(t)})\right)-\frac{L\ln 2}{\delta_2 \omega}\right]\right)\!\!,
	    \end{split}
	\end{equation}   
	where $ \Phi\left(q\right)\triangleq\frac{1}{\sqrt{2\pi}}\int_{q}^{\infty}\exp(-\frac{u^2}{2})du $,  $ \Gamma^{(t)}_i\triangleq1-\frac{1}{(1+\gamma_{i,\pi_{m}^{(t)}}^{(t)}(\bm{\pi}^{(t)},\mbox{\textbf{p}}^{(t)}))^2} $ is   the channel
	dispersion, $ \gamma_{i,\pi_{m}^{(t)}}^{(t)} $ is the SINR at vehicle $i$ at time slot $t$},   $ \omega $  is the bandwidth of the channel, and $ \delta_2\triangleq\delta-\delta_1 $ is the information transmission time,  with $ \delta_1 $ as the dedicated time to acquire the vehicles' angular parameters (i.e., location and speed). {\color{black} It is worth noting that according to the SIC mechanism, for a given decoding order decision $\bm{\pi}^{(t)}$, a vehicle can correctly estimate the   $ \pi_{l'}^{(t)} $-th  message if all the previous messages  $ \pi_{m}^{(t)} $ $ \forall 1 \leq m \leq  l'-1$ have been correctly estimated and removed regardless of whether  this vehicle is interested of previous   $ \pi_{m}^{(t)} $ $ \forall 1 \leq m \leq  l'-1$ messages or not. 
}

}

%%%%%%%%%%%%%%%%%%%%%%%%%%%%%%%%%%%%%%%%%%%%%%%%%%%%%%%%%%%%%%%

\subsection{Age of Information}
The instantaneous AoI of the $ \pi_{l'}^{(t)} $-th physical process  at vehicle $ i $  evolves according to 
\begin{equation}\label{Ins_AoI}
	\Delta_{i\pi_{l'}}^{(t)}\!(\bm{\pi}^{(t)}\!,\!\mbox{\textbf{p}}^{(t)}\!)\!=\!\! \begin{cases} \delta,&\mbox{if}~ \mbox{\eqref{Condd} is satisfied},\\ 
		\Delta_{i\pi_{l'}}^{(t-1)}\!+\!\delta,& \mbox{otherwise}.'' \end{cases}
\end{equation}

The time-average  AoI of $f_l$ at vehicle  $ i $ over $ T $ time slots is
	 $  \bar{\Delta}_{{i,l}}(\bm{\pi}^{(t)},\mbox{\textbf{p}}^{(t)})\triangleq\mathbb{E}_{T}\!\!\left[\Delta_{i,l}^{(t)}(\bm{\pi}^{(t)},\mbox{\textbf{p}}^{(t)})\right]=$ $ \frac{1}{T}\sum_{t=1}^{T}	\Delta_{i,l}^{(t)}(\bm{\pi}^{(t)},\mbox{\textbf{p}}^{(t)}) $. Consequently, the total time-average  AoI    can be expressed as follows:
	 \begin{equation}\label{Avv_AoI}
	 	\begin{split}
	 	\bar{\Delta}(\bm{\pi}^{(t)},\mbox{\textbf{p}}^{(t)})&=\sum_{i=1}^{V}\sum_{l=1}^{F} r_{i,l}\bar{\Delta}_{{i,l}}(\bm{\pi}^{(t)},\mbox{\textbf{p}}^{(t)})\\
	 	&=\frac{1}{T}\sum_{i=1}^{V}\sum_{t=1}^{T}\sum_{l=1}^{F} r_{i,l}	\Delta_{i,l}^{(t)}(\bm{\pi}^{(t)},\mbox{\textbf{p}}^{(t)}). 
	 	\end{split}
	 \end{equation}
Note that the maximum value of $ \bar{\Delta}_{{i,l}} $ is $ \delta(T+1)/2 $, which corresponds the case of no update about $f_l$ is received at vehicle $ i $ over the $ T $ time slots. Thus, the maximum value (upper bound) of the total time-average  AoI $ \bar{\Delta}^{\mbox{\scriptsize max}} $ can be expressed as follows:
\begin{equation}\label{max_AoI}
	\bar{\Delta}^{\mbox{\scriptsize max}} = \frac{\delta (T+1)}{2}\sum_{i=1}^{V}\sum_{l=1}^{F}r_{i,l},
\end{equation}   
which corresponds the case of no update is received by any vehicle during the $ T $ time slots. The minimum value (lower bound) of the total time-average  AoI $ \bar{\Delta}^{\mbox{\scriptsize min}} $ corresponds the case of  each vehicle is able to decode its required messages in each time slot. Consequently, $ \bar{\Delta}^{\mbox{\scriptsize min}} $
  can be expressed as:   
\begin{equation}\label{min_AoI}
	\bar{\Delta}^{\mbox{\scriptsize min}} =\delta \sum_{i=1}^{V}\sum_{l=1}^{F}r_{i,l}.
\end{equation}   

\subsection{A Toy Example}

Fig. \ref{SysModel} illustrates a schematic   diagram of the considered system model with $F=4$ processes and $V=4$ vehicles, the vehicles' information demand matrix is  
\begin{equation}\label{demR}
	\mathbf{R}=[r_{i,l}]_{\footnotesize{ 4\times 4}}=\begin{blockarray}{ccccc}
		f_1 & f_2 & f_3 & f_4  \\
		\begin{block}{[cccc]c}
			1 & 1 & 0 & 0 & v_1\\
			0 & 0 & 1 & 1 &  v_2 \\
			0 & 1 & 0 & 1 &  v_3 \\
			1 & 0 & 0 & 0 & v_4 \\
		\end{block}
	\end{blockarray}.
\end{equation}
The messages are superposed at the RSU according to a dummy  decoding order decision $\bm{\pi}^{(t)} = [1, 2, 3, 4]$. Let us assume that the power is properly allocated, the SIC at the vehicles  is performed as follows. {At a given vehicle $v_i$, the message   $f^{(t)}_1$ can be obtained by applying direct decoding from the received signal while   treating the other messages   as interference.  If the condition in \eqref{Condd} is satisfied and $f^{(t)}_1$ is correctly decoded,  the AoI of process $f_1$ at   vehicle $v_i$ will be updated if $r_{i,1}=1$ (i.e., process $f_1$ is of interest to $v_i$).  The SIC procedure is performed to get  message   $f^{(t)}_2$ if the previous message   $f^{(t)}_1$ was correctly decoded, such that       $f^{(t)}_1$ is removed and   $f^{(t)}_3$ and $f^{(t)}_4$ are treated as interference, and so on.} It is worth noting that the iterations of the SIC at each vehicle depends on its requirements and the decoding order decision $\bm{\pi}^{(t)}$. For example, in Fig. \ref{SysModel} vehicle  $v_1$ applies direct decoding to get $f^{(t)}_1$ and one SIC iteration to get $f^{(t)}_2$, vehicles  $v_2$ and $v_3$ apply direct decoding followed by three SIC iterations    to get $f^{(t)}_4$, and vehicle  $v_4$ applies only the direct decoding to get $f^{(t)}_1$. 

It is also worth mentioning that, unlike the conventional NOMA scenarios where the messages can be superposed   according to the ranking of the communication channel conditions, both the power allocation and the decoding order decision $\bm{\pi}^{(t)}$ should be optimized to consider the vehicles demands and the value of the AoI at each vehicle.

\begin{figure}[ht]
	\begin{center}
		\includegraphics[width=1\linewidth]{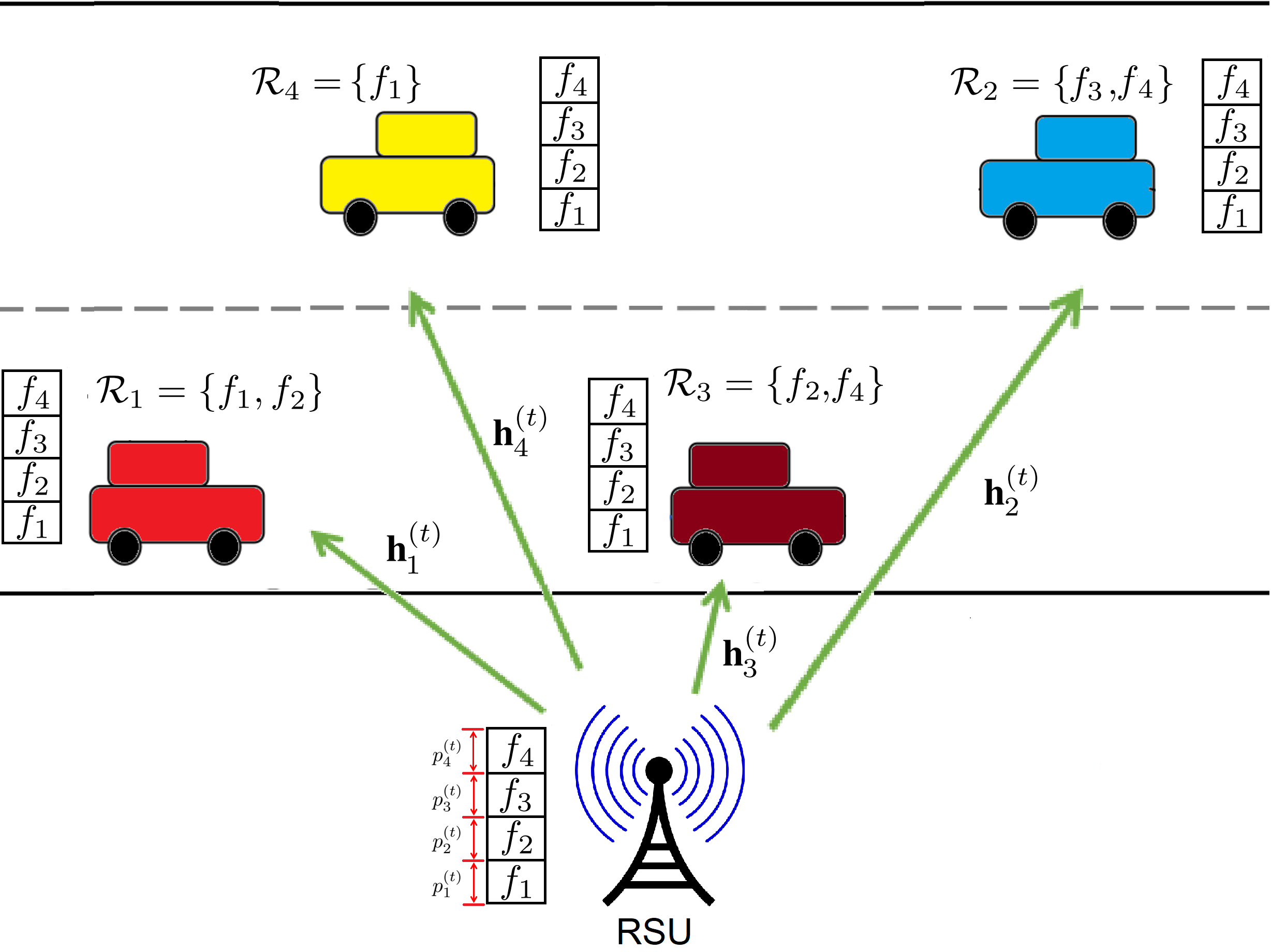}
		\caption{System model with $V=4$ vehicles,  $F=4$ processes, information demand $\mathbf{R}$ in \eqref{demR}, and a   decoding order decision $\bm{\pi}^{(t)} = [1, 2, 3, 4]$.}\label{SysModel}
	\end{center} 
\end{figure}

\section{Multi-objective Problem Statement}\label{Prob}
We consider minimizing the time-average AoI of each process at the vehicles $ \bar{\Delta}(\bm{\pi}^{(t)},\mbox{\textbf{p}}^{(t)}) $  as well as the time-average   power consumption at RSU, $ \bar{{p}}^{(t)} = \frac{1}{T}\sum_{t=1}^{T}\sum_{l=1}^{F}p^{(t)}_l $.  The   multi-objective  problem is thus formulated  as follows: 
     
\begin{subequations}\label{OP1}
	\begin{alignat}{2}
		\mbox{\textbf{P1}}~	&\min_{\bm{\pi}^{(t)},\mbox{\textbf{p}}^{(t)}}&& \left\{\bar{\Delta}(\bm{\pi}^{(t)}\!,\mbox{\textbf{p}}^{(t)}\!), \bar{{p}}^{(t)}\right\} \label{OPP1}\\
		&\mbox{s.t.}&& {\sum_{l=1}^{F}p_l^{(t)} \leq P^{\mbox{\scriptsize max}}},\label{con11}\\
					&&~~   & p_l^{(t)}\geq 0, ~ \forall~  1 \leq l \leq  F,\label{con22}\\
		&                  & ~~     &  \pi_{l'}^{(t)} \in \mathcal{F},  ~ \forall~  1 \leq l' \leq  F, \label{con3}\\
			&                  & ~~     &  \mbox{card}(\bm{\pi}^{(t)}) \leq F. \label{con4}
	\end{alignat}
\end{subequations}
Constraints \eqref{con11} and \eqref{con22}  guarantee that the allocated power is positive and  less than the maximum transmission power of the RSU. Constraint \eqref{con3} guarantees that the decoding order decision captures only  the physical processes  and \eqref{con4}   guarantees there is no repeated processes, with $\mbox{card}(\cdot)$ as a cardinality operator.

%\subsection{Problem Decomposition Strategy}  
% Decomposition strategy is an effective approach
% to design a multi-objective optimization algorithm, in which the problem is decomposed into $ J $ scalar optimization
% sub-problems by $ J $ objective-preference weights each weight corresponds to a particular sub-problem \cite{9022866}.  In recent DRL approaches for multi-objective optimization, the
% decomposition is performed as a weighted sum approach
% which considers the linear combination of different objectives with a set of uniformly spread weights \cite{9714721}. 
% Solving each sub-problem    leads to a
%   potential Pareto optimal point, the   desired Pareto front can
% be obtained as the set of   the $J$ Pareto optimal points.
% 

Keeping in mind   the trade-off between   these two objectives and the fact that   they  have different units, ranges, and orders of magnitude, they
should be normalized such that they have similar ranges \cite{marler2004survey}. The common strategy to handle multi-objective optimization is to convert
the problem into a single objective optimization problem, which can be implemented using three approaches. The first approach referred to as 
$ \epsilon $-constraint approach, in which one   objective
is selected to be the primary objective and the others
appear as constraints with respect to the  auxiliary parameter $ \epsilon $ \cite{9131855}. This approach has three major shortcomings: (i) The value of the  auxiliary parameter $ \epsilon $ should be carefully adjusted as it may affect the problem feasibility; (ii) The selection of the primary objective function; (iii)  More than one auxiliary parameter may be required to constrain    objectives of different ranges.  The second approach is   Tchebycheff approach, in which the primary objective is to optimize (maximize or minimize) an auxiliary parameter  and all  objectives appear as  weighted constraints with respect to the  auxiliary parameter \cite{9485101}. The third approach is the weighted sum, in which    the objectives are combined into a single function using
prefixed weights \cite{kim2006adaptive,marler2004survey,augusto2012new}. 

The weighted-sum is an effective approach
 to handle a multi-objective optimization problem,  where each weight corresponds to a particular sub-problem. The solutions   of $J$ sub-problems constitute a set of the Pareto optimal solutions or the Pareto optimal front \cite{9022866}.

$ \!\!\!\!\!\! $\textbf{Definition 1.} \textit{Let $ \mathcal{X} $ represents  the feasible space of
	the optimization problem}
\begin{equation*}
	\min_{\mbox{\textbf{x}}} \mbox{\textbf{o}}(\mbox{\textbf{x}})=\left\{o_1(\mbox{\textbf{x}}), o_2(\mbox{\textbf{x}}), \cdots, o_K(\mbox{\textbf{x}})\right\} ~~ \mbox{subject to: }~ \mbox{\textbf{x}} \in \mathcal{X}
\end{equation*}
\textit{A point  $ \mbox{\textbf{x}}^* \in \mathcal{X} $   is a Pareto optimal point  if and only if  there does not exist another point, $ \mbox{\textbf{x}} \in \mathcal{X} $,  such that $ \mbox{\textbf{o}}(\mbox{\textbf{x}}) \leq \mbox{\textbf{o}}(\mbox{\textbf{x}}^*) $, and $ o_k(\mbox{\textbf{x}}) < o_k(\mbox{\textbf{x}}^*) $ for at least one objective function. The set of all Pareto optimal
	points is called the Pareto frontier.}
   
To obtain the Pareto optimal fronts for \eqref{OP1}, we consider the normalized 
weighted metric method, which entails
minimizing the difference between the objectives and the corresponding
  utopia solutions. Fortunately, we can obtain both  the Utopia and Nadir solutions of the two objectives. The Utopia and Nadir solutions of time-average AoI  are $ \bar{\Delta}^{\mbox{\scriptsize max}} $ and $ \bar{\Delta}^{\mbox{\scriptsize min}} $  as expressed in \eqref{max_AoI} and \eqref{min_AoI}, respectively. While for the  time-average power consumption, the Nadir and Utopia    solutions are the maximum transmission power of the RSU $ P^{\mbox{\scriptsize max}} $  and   the minimum transmission power of the RSU ($ P^{\mbox{\scriptsize min}}=0 $), respectively.    
  Consequently,  we define   the objective function as:
 
 \begin{equation}\label{Obj}
 	\begin{split}
 		{O}(\bm{\pi}^{(t)}\!,\mbox{\textbf{p}}^{(t)})\!=& \zeta\frac{\bar{\Delta}(\bm{\pi}^{(t)},\mbox{\textbf{p}}^{(t)}) \!-\!\bar{\Delta}^{\mbox{\scriptsize min}}}{\bar{\Delta}^{\mbox{\scriptsize max}}-\bar{\Delta}^{\mbox{\scriptsize min}}}\!+\!(1\!-\!\zeta)\frac{\bar{{p}}^{(t)}\!-\!P^{\mbox{\scriptsize min}}}{P^{\mbox{\scriptsize max}}\!-\!P^{\mbox{\scriptsize min}}},
 	\end{split}
 \end{equation}
where $ 0 \leq \zeta \leq 1$ is a relative weight to give preference to minimize the AoI or the power.
It is worth noting that the objective function in  \eqref{Obj}  is dimensionless, bounded by $[0, 1]$, and the decision-maker can set the   
 objective-preference weight to any desired value  $ 0 \leq \zeta \leq 1$. This is not the case in other multi-objective optimization methods, such as  $ \epsilon $-constraint method, in which the value of $ \epsilon $ should be selected carefully to avoid 
 infeasibility issues. 
 
 Henceforth,  the multi-objective
 optimization problem $ \mbox{\textbf{P1}} $ can be transformed as follows:
  
 \begin{subequations}\label{P2}
 	\begin{alignat}{2}
 		\mbox{\textbf{P2}}~	&\min_{\bm{\pi}^{(t)},\mbox{\textbf{p}}^{(t)}}&& {O}(\bm{\pi}^{(t)}\!,\mbox{\textbf{p}}^{(t)}) \label{OPP122}\\
 		&\mbox{s.t.}&&  \eqref{con11}-\eqref{con4}.
 	\end{alignat}
 \end{subequations}
 
 The optimization problem in \eqref{P2} is a mixed-integer non-linear programming (MINLP) problem that involves    discrete (decoding order decision $ \bm{\pi}^{(t)}$) and continuous (power allocation  $ \mbox{\textbf{p}}^{(t)}$)
 decision making over multiple transmission time slots. Conventional optimization methods are not appropriate for solving the problem  \eqref{P2} which evolves over time, hence we resort to the DRL strategies for solving problem \eqref{P2}.

\section{Hybrid DQN-DDPG -Based DRL Solution}\label{DRLApproach}

In this   section,    a   hybrid DQN-DDPG  DRL model is introduced to solve the   optimization problem  in \eqref{P2} for a given value of  the objective-preference weight $ \zeta $.

\subsection{Theoretical Preliminaries}
Reinforcement learning is the process of
learning    by an agent to maximize the discounted reward over the learning time horizon by interacting with an environment.  
 At each learning epoch, given the current state of the environment, an action is performed by the agent on the environment which transits to the next state and returns  an immediate reward.
 A key metric in training the agent is the $Q$-function  which estimates    the future reward of taking an action  $ \bm{a}$  at a given  state $ \bm{s} $.  It has been shown that learning based on $Q$-function  (also referred to as $ Q $-learning)   converges towards an   optimal
 solution after visiting each state-action pair   with sufficient number of learning iterations \cite{sutton2018reinforcement}. Such a learning approach is impractical in the following   scenarios: (1) The  number of   state-action pairs is very large; (2) The action  space and/or state space are/is continuous. Utilizing deep neural networks (DNN)   to approximate the $Q$-function is referred to as DRL, in which the $Q$-function is written as $ Q(\bm{s},\bm{a}\mid \bm{\theta}) $, where $ \bm{\theta} $ represents the weight vector of the DNN \cite{mnih2015human}. 
Actor-critic networks comprising of  two deep  $ Q $-learning networks  are able to deal with   environments  that have continuous   action spaces \cite{ap2016continuous}. DDPG is an actor-critic algorithm involves two networks, namely, the actor and critic networks. The actor  network     learns   to obtain the best action at a given state, while the critic      evaluates
the reward of the state-action pair \cite{ap2016continuous}.

{ \color{black} It is worth mentioning  that the DRL model is commonly formulated as a Markov
	Decision Process (MDP) problem.  
		An MDP is represented by a tuple $\{\mathcal{S}, \mathcal{A}, \vartheta, \rho\}$, where $\mathcal{S}$ is the state space that consists of the set of
	all possible   states, $\mathcal{A}$ is a finite set of actions
	from which the agent can choose, $\vartheta: \mathcal{S}\times\mathcal{A}\times\mathcal{S} \rightarrow [0,1]$   is a transition probability   which defines the
	probability of observing a state  after executing an action at a given environment's state, and  $\rho: \mathcal{S}\times\mathcal{A}\rightarrow\mathbb{R}$  is the expected reward of performing an action at a given state. {\color{black} We note that  discretizing  continuous action space  increases the problem’s dimensionality. 
		Thus, DDPG   \cite{ap2016continuous} is applied to handle continuous action space.
		The DDPG algorithm maintains a parameterized actor policy,  which    maps states to a probability distribution over
		the actions \cite{ap2016continuous}.  
		In this context, a   hybrid DQN-DDPG  DRL model is developed such that the DQN handles the decoding order decision (discrete value) and the actor-critic DDPG handles the power allocation (continuous value) decision.
} The following section defines the state space, action space, and reward of the designed DRL model.

\subsection{DRL Model} \label{HDQNDDPG}
Our goal is to design a DRL system that jointly optimizes 
the decoding order decision (discrete value) and the power allocation (continuous value) decision to minimize the objective function in \eqref{P2}.    In this context, we
develop a DRL model   that involves the definition of the environment state, the action, and the    immediate reward function $ \rho^{(t)} $ as follows:  

\subsubsection{The Environment State Space} 
The state space  is denoted as $ \mathcal{S} $, in which the state  at time slot $ t $ captures  the  communication    channel between the RSU and the vehicles as well as the   AoI of the required processes at the vehicles. Thus, the   state at time slot $ t $   is given by

\begin{equation}
	\bm{s}^{(t)}=\left\{\{\sum_{i=1}^{V}r_{i,l}  \chi^{(t)}_i\}, \{\sum_{i=1}^{V}r_{i,l} \Delta_{i,l}^{(t)} \}\right\}_{l=1}^F.
\end{equation}

\subsubsection{Action  Space} 
At each decision-making instant, based on the observed state,   the DQN and actor-critic models make  an action  to order the physical process for encoding and to allocate the power, respectively. Under this setup, the action space $ \mathcal{A}  $ represents pairs of actions $ \mbox{\textbf{a}}^{(t)} =(\mbox{\textbf{a}}_{\bm{\pi}}^{(t)}, \mbox{\textbf{a}}_{\bm{p}}^{(t)}) $, such that

\begin{itemize}
     \item $ \mbox{\textbf{a}}_{\bm{\pi}}^{(t)}=[\pi^{(t)}_1, \pi^{(t)}_2, \cdots, \pi^{(t)}_{F}] $ is the decoding order decision which is obtained using the DQN agent.  
     \item $ \mbox{\textbf{a}}_{\bm{p}}^{(t)}=  [\alpha_1^{(t)}, \alpha_2^{(t)}, \cdot, \alpha_F^{(t)}]P^{max} $ is the power allocation decision, with      $ \bm{\alpha}^{(t)}=[\alpha_1^{(t)}, \alpha_2^{(t)}, \cdot, \alpha_F^{(t)}] $ as the output of the actor-critic/DDPG agent.
\end{itemize}

\subsubsection{Reward} 
Based on the observed state and the agents' action, 
an immediate reward is returned by the environment that reflects the suitability of the action  to minimize the objective function with a given value of the objective preference weight.   For the considered optimization problem,   a
good decision on process decoding order   and power allocation  can minimize the AoI and power consumption with the given preference weight $ \zeta $. To
reflect the  quality of  the action   taken by the agents, the immediate
reward at time slot $ t $ is expressed as:
\begin{equation}\label{Rew}
\begin{split}
	\rho^{(t)}\!\!=& \exp\left(-\zeta\frac{\bar{\Delta}_t(\bm{\pi}^{(t)},\mbox{\textbf{p}}^{(t)}) \!-\!\bar{\Delta}^{\mbox{\scriptsize min}}}{\bar{\Delta}_t^{\mbox{\scriptsize max}}-\bar{\Delta}^{\mbox{\scriptsize min}}}\!-\!(1\!-\zeta)\frac{\hat{p}^{(t)}-P^{\mbox{\scriptsize min}}}{P^{\mbox{\scriptsize max}}-P^{\mbox{\scriptsize min}}}\right)\\
	&- \Upsilon^{(t)},
	\end{split}
\end{equation}
where $ \Upsilon^{(t)}=\kappa[ \sum_{l=1}^F\alpha_l^{(t)}-1]^+ $ is  a penalty function  with $ [x]^+ =\max\{x,0\} $ and $\kappa$ as a penalty constant, $ \hat{p}^{(t)}=\frac{1}{t}\sum_{t'=1}^{t}\sum_{l=1}^{F} {p_l}^{(t')} $, $\bar{\Delta}_t(\bm{\pi}^{(t)},\mbox{\textbf{p}}^{(t)})$, and  $ \bar{\Delta}_t^{\mbox{\scriptsize max}} $,    are obtained by replacing $T$ by $t$ in \eqref{Avv_AoI} and \eqref{max_AoI},   respectively.} 
{ It is worth noting that 
the parameters in equation \eqref{Obj} are  redefined in \eqref{Rew} to represent the AoI and power consumption at time instant $t$. Keeping in mind that our aim is to minimize the objective function, while the DRL learns to maximize the accumulative reward,    the exponential function is chosen to define the immediate reward which stabilizes the offline training \cite{8954939}. Finally, $ \Upsilon^{(t)} $ represents a penalty of violating the constraint \eqref{con11}.}

%The DRL model 
%  aims to find an optimal policy
%that maximizes the expected long-term accumulative discounted reward, which is expressed as 
%\begin{equation}
%     \varphi^{(t)}= \sum_{\iota=0}^{\infty} \varpi_\iota	\rho^{(t+\iota)},
%\end{equation}
%where  $ \varpi_\iota \in [0,1] $ is   the discount factor, which is introduced to 
%balance the immediate  and future cumulative rewards. 

\subsection{Hybrid DQN-DDPG Algorithm}
A schematic block diagram of the proposed hybrid DQN-DDPG DRL model is shown in Fig. \ref{FigHRL}, which
illustrates the action-making agents for decoding order and
power allocation as well as the learning process for both agents for a given value of the objective-preference weight $ \zeta $.  
The training algorithm of the hybrid DQN-DDPG model is illustrated in \textbf{Algorithm~1}.  The  objective-preference weight $ \zeta $, and    DQN weights $\theta^Q $, critic   network weights $\theta^{Q_c} $,    actor network weights $\theta^\mu $ are the input of the algorithm. For each training episode, the environment is initialized and an initial state is obtained (line \ref{l4}). At every time step $ t $, the  DQN model obtains a decoding order action $ \bm{a}_{\bm{\pi}}^{(t)} $ using the $ \bar{\epsilon} $-greedy policy   to balance the exploitation of known actions and the  exploration of new
actions (line \ref{l5}). The DDPG model obtains a power allocation action $ \bm{a}_{\bm{P}}^{(t)} $ using the actor network $ \mu(\bm{s}^{(t)}\mid \theta^\mu) $ (lines \ref{l6}-\ref{l7}). 
A combined action $ \bm{a}^{(t)} =(\bm{a}_{\bm{\pi}}^{(t)}, \bm{a}_{\bm{P}}^{(t)}) $ is applied on 
the environment 
which returns an immediate reward $ \rho^{(t)} $ and   transits to a new state  $ \bm{s}^{({t+1})} $ (line \ref{l8}). The   transition tuple $ \{\bm{s}^{({t})},\bm{a}^{(t)}, \rho^{(t)},\bm{s}^{({t+1})}\}$ is added to the replay buffer  \textbf{B} (line \ref{l9}).  A mini-batch of   transitions are sampled from the replay buffer to update the deep networks (line \ref{l11}). 

 \begin{figure}[!ht]
		\begin{center}				\includegraphics[width=.8\linewidth]{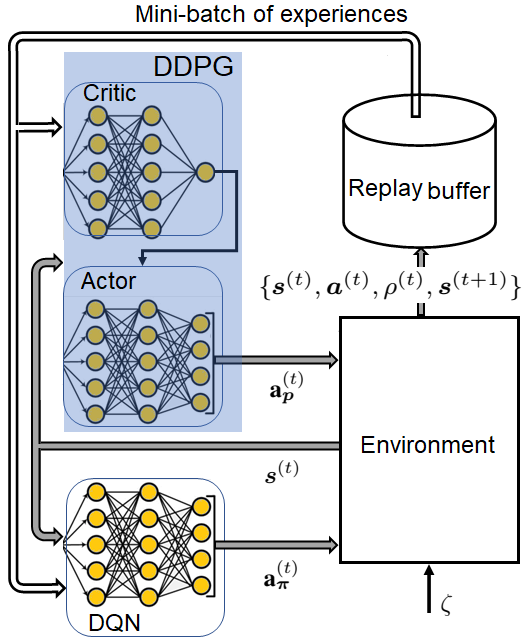}
				\caption{Hybrid DQN-DDPG model   for age-optimum information dissemination.} \label{FigHRL}
			\end{center}
	\end{figure}

\begin{algorithm}[!ht]     
	\caption{\footnotesize  Hybrid DQN-DDPG training algorithm   for age-optimum information dissemination.}\label{DeepLearning}
	\begin{algorithmic}[1]\footnotesize
		\State \textbf{Input:} Objective-preference weight $ \zeta $, No. of episodes,    DQL network weights $\theta^Q $, critic   network weights $\theta^{Q_c} $, and   actor network weights $\theta^\mu $;\label{2}
		\State {\color{black}Initialize target network $\theta^{Q'} \leftarrow\theta^Q $,  $\theta^{Q'_c}\leftarrow\theta^{Q_c} $, and    $\theta^{\mu'}\leftarrow\theta^\mu $;}
		\State  \textbf{For $ episode=1 $ to} No. of episodes \textbf{do}\label{l3} 
		\State  Initialize the
		environment and receive the initial state $ \bm{s}^{(1)} $; \label{l4}
		\State $~$\textbf{Repeat:} 
		\State $ ~~$With probability $ \bar{\epsilon} $, select randomly an action $\bm{a}^{(t)}_{\bm{\pi}} \in \mathcal{A}_{\bm{\pi}}$; Otherwise, select
		 $ \bm{a}_{\bm{\pi}}^{(t)}=\arg\max\limits_{\forall \bm{a}_{\bm{\pi}}^{(t)}\in \bm{\mathcal{A}}_{\bm{\pi}}}Q(\bm{s}^{(t)},\bm{a}^{(t)}_{\bm{\pi}}\mid\bm{\theta}^Q) $; \label{l5}
		 \State $ ~~$ Obtain   $ \bm{\alpha}^{(t)}=[\alpha_1^{(t)}, \alpha_2^{(t)}, \cdots, \alpha_F^{(t)}]\triangleq \mu(\bm{s}^{(t)}\mid \theta^\mu){\color{black}+N}$, {\color{black}where $N$ is the exploration noise \cite{ap2016continuous};}\label{l6}
	 		  \State $ ~~$ Obtain the power allocation action $ \bm{a}_{\bm{P}}^{(t)}= \bm{\alpha}^{(t)} P^{max} $; \label{l7}
		\State $ ~~$  Observe the reward  $ \rho^{(t)} $ using \eqref{Rew} and    next state $\bm{s}^{({t+1})}  $ by applying the action $ \bm{a}^{(t)} =(\bm{a}_{\bm{\pi}}^{(t)}, \bm{a}_{\bm{P}}^{(t)}) $; \label{l8} 
		\State $ ~~$	Store the transition $ \{\bm{s}^{({t})},\bm{a}^{(t)}, \rho^{(t)},\bm{s}^{({t+1})}\}$ in \textbf{B};
		$ t=t+1 $;\label{l9}
		\State	$~$\textbf{Until} terminal state $t=T$;\label{20}
		\State $ ~~~$ 
		Sample a   mini-batch of $ M $ transitions from \textbf{B};	\label{l11}			
		\State $ ~~~$  Update the weight   $\theta^Q $,   $\theta^{Q_c} $ and   $\theta^\mu $ by minimizing the  corresponding loss functions in \eqref{Loss_Funs};\label{l12}
		\State	{\color{black}Update the target networks
		\begin{equation*}
			\begin{split}
&\theta^{Q'}\leftarrow \tau \theta^{Q}+(1-\tau)\theta^{Q'},\\
&\theta^{Q'_c}\leftarrow \tau \theta^{Q_c}+(1-\tau)\theta^{Q'_c},\\
&\theta^{\mu'}\leftarrow \tau \theta^{\mu}+(1-\tau)\theta^{\mu'}.
			\end{split}
		\end{equation*}}
		\State \textbf{End for}     
		\normalsize
	\end{algorithmic}
\end{algorithm}

The \textit{Adam} optimizer    \cite{kingma2014adam} is considered    to update the weights of the DNNs with an aim to minimize the following loss functions

\begin{equation}\label{Loss_Funs}
\begin{split}
 &L({\theta}^Q)\! =\!\sum_\iota( \rho^{(\iota)}\! +\!\! \max_{\forall \bm{a}_{\bm{\pi}}^{'} \in \mathcal{A}}\!Q'(\bm{s}^{'},\bm{a}_{\bm{\pi}}^{'},{\theta}^{Q'})\!-\!Q(\bm{s}^{(\iota)},\bm{a}_{\bm{\pi}}^{(\iota)},{\theta}^Q))^2,\\
 &L({\theta}^{Q_c})  =\sum_\iota( {\color{black}y^{(\iota)}}  -Q(\bm{s}^{(\iota)},\bm{a}_{\bm{P}}^{(\iota)},{\theta}^{Q_c}))^2,  \\
 & L({\theta}^\mu) = -Q(\bm{s}^{(\iota)},\bm{a}_{\bm{P}}^{(\iota)},{\theta}^{Q_c},{\theta}^{\mu}),
\end{split} 
\end{equation}
{\color{black}where $y^{(\iota)}=\rho^{(\iota)} +Q_c'(\bm{s}^{'},\bm{a}_{\bm{P}}^{'},{\theta}^{Q_c'})$ \cite{ap2016continuous}.}

{\color{black} It is worth noting that both the DQN and DDPG agents work jointly such that they observe the same observation or environment state and the shared reward is not sparse and its value depends on the actions of both agents, which prevents a lazy agent situation. In other words, let us assume that the DQN model tends to become a lazy agent. In such a case,  the reward of  DQN model   will decrease and the DDPG will not be able to compensate this behavior, i.e., cannot force the  DQN model to learn a good policy. The same applies if the DDPG model tends to become a lazy agent, then it will not take advantage of the successful actions of DQN. That is, if the action of the DDPG is bad the reward will decrease regardless of how good is the action of DQN model. Such a joint effect of the actions of the two agents on the value of the reward and the shared environment observation prevent both agents from becoming a lazy agent, {\color{black} which may arise due to partial observability \cite{sunehag2017value, sunehag2018value} or sparse reward scenarios \cite{liu2023lazy}.''}}

\section{Meta Reinforcement Learning}\label{METADRL}
In Section \ref{DRLApproach}, a hybrid DQN-DDPG  DRL model has been developed
to solve the age-optimum information dissemination in vehicular networks problem. However, it is worth noting that  a predetermined objective-preference weight
 is required before training the model. Consequently,  the quality of
the inferred solution   depends   on whether the
objective-preference weight has been observed during the training stage. 
If  an unseen objective-preference weight is encountered, a new model should
  be trained from scratch  which is inefficient approach to construct the Pareto front as it requires a large
  number of models   to be trained and stored.  Inspired by 
  recently proposed algorithms for    fast adaptation in DRL models\cite{finn2017model, 9714721}, this section introduces a  meta reinforcement
   learning approach  to increase the diversity and quality    of the
     solutions. {\color{black} 	The training of the meta RL is performed based on the  multi step model-agnostic meta-learning (MAML) approach \cite{ji2022theoretical},   which consists 
     	two nested stages. The inner stage
     	performs   multiple     gradient descent steps to update the deep notworks parameters  for  a given value   of the  relative weight, while the outer stage   enables the updating of the deep notworks' parameters     over all the sampled values of relative weight.
     }

 \subsection{Meta Reinforcement Algorithm}\label{Meta}
Training the meta-DRL model is an important part of the meta-based algorithm as it     yields the parameters for the neural
network that can   adapt quickly to a new task. The goal of meta-learning is to ensure that the meta-based model is capable of optimizing the
objective function with any objective-preference weight   after a small number of
fine-tuning updates. {\color{black} 
Suppose a set $ J $  tasks\footnote{Each  task is a scalar optimization
sub-problem with an objective-preference weight} are available for learning and the corresponding objective-preference weights    are sampled
based on a probability distribution $ \bm{\Lambda} $, the $j$-th task is
associated with    loss functions $ L_j(\tilde{\theta}^Q) $, $ L_j(\tilde{\theta}^{Q_c}) $, and  $ L_j(\tilde{\theta}^\mu) $     parameterized by the DQL network weights $\tilde{\theta}^Q $, critic   network weights $\tilde{\theta}^{Q_c} $, and   actor network weights $\tilde{\theta}^\mu $, respectively.   MAML approach   guarantees convergence \cite{ji2022theoretical}, which aims at finding   good initial parameters ($\tilde{\theta}^{Q^*} $,   $\tilde{\theta}^{Q_c^*} $,  $\tilde{\theta}^{\mu^*} $) 
such that after
observing a new task, a few gradient descend steps (fine-tuning steps) starting from such initial parameters can
efficiently approach the optimizer  of the corresponding loss functions. {The 
	multi-step MAML consists of two nested stages,  the inner stage
	performs   multiple     gradient descent steps for each individual task, while the   meta parameters over all the sampled tasks
	are updated using the outer stage. These two nested stages provide diverse and representative training, which  mitigates the bias in  the training \cite{ji2022theoretical}.} Consequently, 
    the inner stage of each task initializes at the meta parameters, (i.e., $ \tilde{\tilde{\theta}}^Q_{j_{0}}=\tilde{{\theta}}^Q $, $ \tilde{\tilde{\theta}}^{Q_c}_{j_{0}}={\tilde{\theta}}^{Q_c} $, $ \tilde{\tilde{\theta}}^{\mu}_{j_{0}}={\tilde{\theta}}^{\mu} $) and  runs  $ M $
 gradient descent steps as
\begin{equation}
	\begin{aligned}\label{inner_grad}
	&\tilde{\tilde{\theta}}^Q_{j_{k+1}}=\tilde{\tilde{\theta}}^Q_{j_{k}}-\alpha_{Q} \nabla L_j(\tilde{\tilde{\theta}}^Q_{j_{k}}), \\
		&\tilde{\tilde{\theta}}^{Q_c}_{j_{k+1}}=\tilde{\tilde{\theta}}^{Q_c}_{j_{k}}-\alpha_{Q_c} \nabla L_j(\tilde{\tilde{\theta}}^{Q_c}_{j_{k}}),
		\\
		&\tilde{\tilde{\theta}}^{\mu}_{j_{k+1}}=\tilde{\tilde{\theta}}^{\mu}_{j_{k}}-\alpha_{\mu} \nabla L_j(\tilde{\tilde{\theta}}^{\mu}_{j_{k}}),
	\end{aligned}
\end{equation} 
where $ \alpha_{Q} $, $ \alpha_{Q_c} $, and $ \alpha_{\mu} $ represent the inner step size of the  DQL,  
 critic, and   actor networks, respectively. Consequently, the overall meta
 goal is given by
 \begin{equation}\label{meta_obj}
 	\begin{aligned}
 		& \min_{\tilde{{\theta}}^Q} \mathcal{L}(\tilde{{\theta}}^Q)\coloneqq \mathbb{E}_{j\sim \bm{\Lambda}}\left[\mathcal{L}_j(\tilde{{\theta}}^Q)\right]\coloneqq \mathbb{E}_{j\sim \bm{\Lambda}}\left[L_j(\tilde{\tilde{\theta}}^{Q}_{j_{M}}(\tilde{\theta}^{Q}))\right],\\
 		&\min_{\tilde{{\theta}}^{Q_c}} \mathcal{L}(\tilde{{\theta}}^{Q_c})\coloneqq \mathbb{E}_{j\sim \bm{\Lambda}}\left[\mathcal{L}_j(\tilde{{\theta}}^{Q_c})\right]\coloneqq \mathbb{E}_{j\sim \bm{\Lambda}}\left[L_j(\tilde{\tilde{\theta}}^{Q_c}_{j_{M}}(\tilde{\theta}^{Q_c}))\right],\\
 		&\min_{\tilde{{\theta}}^{\mu}} \mathcal{L}(\tilde{{\theta}}^{\mu})\coloneqq \mathbb{E}_{j\sim \bm{\Lambda}}\left[\mathcal{L}_j(\tilde{{\theta}}^{\mu})\right]\coloneqq \mathbb{E}_{j\sim \bm{\Lambda}}\left[L_j(\tilde{\tilde{\theta}}^{\mu}_{j_{M}}(\tilde{\theta}^{\mu}))\right].
 	\end{aligned}
 \end{equation} 
Then the outer stage of meta update is  gradient decent steps to minimize \eqref{meta_obj}. In \cite{ji2022theoretical}, a simplified form of gradient of the losses has been derived using the chain rule, which can be expressed as 
\begin{equation} 
	\begin{aligned}
		&  \nabla\mathcal{L}_j(\tilde{{\theta}}^Q)=\left[\prod_{k=0}^{M-1}(1-\alpha_{Q}\nabla^2L_j(\tilde{\tilde{\theta}}^{Q}_{j_{k}}))\right]\nabla L_j(\tilde{\tilde{\theta}}^{Q}_{j_{M}}),\\
		&\nabla\mathcal{L}_j(\tilde{{\theta}}^{Q_c})=\left[\prod_{k=0}^{M-1}(1-\alpha_{{Q_c}}\nabla^2L_j(\tilde{\tilde{\theta}}^{{Q_c}}_{j_{k}}))\right]\nabla L_j(\tilde{\tilde{\theta}}^{{Q_c}}_{j_{M}}),\\
		& \nabla\mathcal{L}_j(\tilde{{\theta}}^\mu)=\left[\prod_{k=0}^{M-1}(1-\alpha_{\mu}\nabla^2L_j(\tilde{\tilde{\theta}}^{\mu}_{j_{k}}))\right]\nabla L_j(\tilde{\tilde{\theta}}^{\mu}_{j_{M}}),
	\end{aligned}
\end{equation} 
  where $ \nabla L_j(\cdot) $ and $ \nabla^2 L_j(\cdot) $ represent the gradient and Hessian operators of the loss function $ L_j(\cdot) $, respectively. Finally, the full gradient descent step of the outer stage can be expressed as 
  
  \begin{equation}\label{otter_grad}
  	\begin{aligned}
  		&\tilde{{\theta}}^Q_{{k'+1}}=\tilde{{\theta}}^Q_{{k'}}-\beta_{Q}\mathbb{E}_{j\sim \bm{\Lambda}}\nabla\mathcal{L}_{j,k'}(\tilde{{\theta}}^Q), \\
  		&\tilde{{\theta}}^{Q_c}_{{k'+1}}=\tilde{{\theta}}^{Q_c}_{{k'}}-\beta_{{Q_c}}\mathbb{E}_{j\sim \bm{\Lambda}}\nabla\mathcal{L}_{j,k'}(\tilde{{\theta}}^{Q_c}),\\ 
  		&\tilde{{\theta}}^\mu_{{k'+1}}=\tilde{{\theta}}^\mu_{{k'}}-\beta_{\mu}\mathbb{E}_{j\sim \bm{\Lambda}}\nabla\mathcal{L}_{j,k'}(\tilde{{\theta}}^\mu),			
  	\end{aligned}
  \end{equation} 
 where $ \beta_{Q} $, $ \beta_{Q_c} $, and $ \beta_{\mu} $ represent the outer step size of the  DQL,  
 critic, and   actor networks, respectively.

\begin{algorithm}[!ht]     
	\caption{\footnotesize  The meta-learning algorithm  for age-optimum information dissemination.}\label{MetaDeepLearning}
	\begin{algorithmic}[1]\footnotesize
		\State \textbf{Input:} Distribution over the   objective-preference weight $ \bm{\Lambda} $, number of meta-learning iterations,  number of
		sampled tasks $J$;
			\State Initialize the  DQL network weights $\tilde{\theta}^Q $, critic   network weights $\tilde{\theta}^{Q_c} $, and   actor network weights $\tilde{\theta}^\mu $;\label{m2}
		\State  \textbf{For $ k'=0 $ to} No. of   meta-learning iterations \textbf{do}\label{Mml3}
		\State Sample $ J $ tasks by distribution $ \bm{\Lambda} $;
		\State $~$ \textbf{For $ j=1 $ to}  $J$ \textbf{do}\label{Mml4} 
		\State $~~$ \textbf{For $ k=0 $ to}  $M$ \textbf{do}\label{Mml5}
		\State $~~~$ Obtain a training set of transition tuples; 
		\State $~~~$  Update the DRL networks parameters using \eqref{inner_grad}; 
		\State $~~$ \textbf{End for}    
		\State $~$ \textbf{End for}    
		\State  Update the DRL networks parameters using \eqref{otter_grad}; 
		\State   \textbf{End for}  
		\State \textbf{Output}    $\tilde{\theta}^{Q^*} \longleftarrow \tilde{\theta}^{Q}$;   $\tilde{\theta}^{Q_c^*}  \longleftarrow \tilde{\theta}^{Q_c} $;   $\tilde{\theta}^{\mu^*}  \longleftarrow\tilde{\theta}^{\mu^*} $.	\normalsize
	\end{algorithmic}
\end{algorithm}
\begin{algorithm}[!ht]     
	\caption{\footnotesize Fine-tuning algorithm to obtain the Pareto front using the meta-model.}\label{FTMetaDeepLearning}
	\begin{algorithmic}[1]\footnotesize
		\State \textbf{Input:} The      objective-preference weight vector of the Pareto front $ \bm{\zeta}=[\zeta_1, \zeta_2, \cdots, \zeta_{\hat{J}}] $, number of  fine-tuning iterations,  well-trained meta-model  DQL network weights $\tilde{\theta}^{Q^*} $, well-trained meta-model  critic   network weights $\tilde{\theta}^{Q_c^*} $, and well-trained meta-model   actor network weights $\tilde{\theta}^{\mu^*} $;\label{FTm2}
		\State $~$ \textbf{For $ j=1 $ to}    $ \hat{J} $ \textbf{do}\label{FTml3} 
		\State $~~$ $ \theta^Q_j, \theta^{Q_c}_j, \theta^\mu_j   $ $\longleftarrow $ $ \tilde{\theta}^{Q^*}, \tilde{\theta}^{Q_c^*}, \tilde{\theta}^{\mu^*}$; 
		\State $~~$ \textbf{For $ episode=1 $ to} No. of   fine-tuning steps \textbf{do}  
		\State $~~~$ $(\theta^Q_j, \theta^{Q_c}_j, \theta^\mu_j) $ $\longleftarrow $ Hybrid DQN-DDPG$(\zeta_j,\theta^Q_j, \theta^{Q_c}_j, \theta^\mu_j ) $ using Algorithm \ref{DeepLearning};
		\State $~~$  \textbf{End for}
	    \State $~~$	Estimate the $j$-th Pareto point using $\theta^Q_j, \theta^{Q_c}_j, \theta^\mu_j  $, and  $\zeta_j $;  
		\State $~$  \textbf{End for} 
		\State \textbf{Output}    the Pareto points.      
		\normalsize
	\end{algorithmic}
\end{algorithm}

The meta-learning algorithm is illustrated in \textbf{Algorithm~2}, in which at   the $ k $-th inner stage iteration a training set of transition tuples is sampled and utilized to update the meta parameters using  \eqref{inner_grad}. At the  $ k' $-th outer stage iteration, two  training sets of transition tuples are independently sampled and utilized to estimate the gradient and Hessian of the loss and update meta parameters using  \eqref{otter_grad}.
}

\subsection{Fine-tuning and Inference}\label{FineT}
In the meta-training stage, we have learned the initial networks
parameters, which have good generalization ability. 
   Pareto solution for any objective-preference weight can be inferred  from the   well-trained meta-model       after undergoing  a few  fine-tuning steps. \textbf{Algorithm~3} illustrates how to infer  a Pareto frontier of $ \hat{J} $ Pareto points corresponding to a set of objective-preference weights $ \bm{\zeta}=[\zeta_1, \zeta_2, \cdots, \zeta_{\hat{J}}] $.

\section{Simulation Results and Discussions}\label{SimSec}
This section introduces simulation results to    evaluate  the proposed framework and solution approaches.   
\subsection{Simulation Parameters}
{Without loss of generality, the considered vehicular network consists of a segment of  a two-lane road along the x-axis and  the vehicles   move along the
	positive or negative directions of the x-axis. The length of the road segment is 3 km and the width of each lane is 3 m. The RSU is located at $ \{1500,50\} $ m and the vehicles' are initialized randomly on the road in both directions.  The vehicles'
	speeds are     randomly drawn from $U(10, 15)  $ m/s. The initial value of the instantaneous AoI of each process of interest at each vehicle is $ \delta $. 
	Unless
	otherwise stated, the considered     values of the system  parameters  are listed in Table
	\ref{TableResults}.     
	\begin{table}[!ht]
		\caption{Simulation Parameters.}
		\begin{center}
			\begin{adjustbox}{width=.48\textwidth}
				\begin{tabular}{|c|c||c|c||c|c|}
					\hline
					\textbf{Parameter}&\textbf{Value} & \textbf{Parameter}&\textbf{Value}   & \textbf{Parameter}&\textbf{Value}    \\ \hline
					$ P^{\mbox{{\footnotesize max}}} $ 	&   $ 0 $ dB  &  $ c_i $ 	& $\sim U(10, 15)  $ m/s  \cite{9557830}  &   $ \omega $	&   $ 10 $ MHz \cite{9557830} \\ \hline    
					$ f_c $ 	&   $ 3 $ GHz  \cite{9557830}&  $ N $ 	&   $ 64 $  \cite{9557830} &  $ L $ 	&   $128 $ byte  \\ \hline   
					$  \varepsilon^{\mbox{{\footnotesize max}}} $ 	&   $ 10^-6 $ &  $ c_0 $ 	&   $ 2.99\times 10^8$ m/s & $ \sigma^2 $ 	&   $ 0.1  $   \\ \hline   					          
				\end{tabular}
			\end{adjustbox}
		\end{center}
		\label{TableResults}
	\end{table}  
}
{\color{black} 	The  implementation of the DQN and DDPG networks involve  three hidden layers of 512, 256, and 128 neurons, respectively. Each layer is followed with a rectified
	linear unit (ReLU)  activation function. The learning rates of the DQN, actor, and critic networks are $ 0.001$, $0.0001 $, and $0.001$, respectively. The mini-batch size for training the hybrid DQN-DDPG model  is $ 64 $, the  discount factor  is $ 0.5 $, and the   penalty constant is $1$. The number of training episodes is $ 1000 $. For the meta training, the  inner step sizes of the
	DQL, critic, and actor networks are $ \alpha_{Q} = 0.1$, $ \alpha_{Q_c}=0.01 $, and $ \alpha_{\mu}=0.001 $ while the outer step sizes are $ \beta_{Q} = 0.01$, $ \beta_{Q_c}=0.01 $, and $ \beta_{\mu}=0.001 $, respectively. The number of meta learning episodes is $500$ and the number of   gradient descent steps and fine-tuning steps is $50$. Fig. \ref{figTraining} shows the  learning curves for the hybrid DQN-DDPG DRL model versus the number of training episodes. It can be noticed that {\color{black} the reward gradually increases, and its average value saturates after sufficient training episodes without a major drop in the average reward. This indicates that the adopted training parameters provide stable learning without over-fitting.}
}

\begin{figure}[ht]
	\begin{center}
		\includegraphics[width=.99\linewidth]{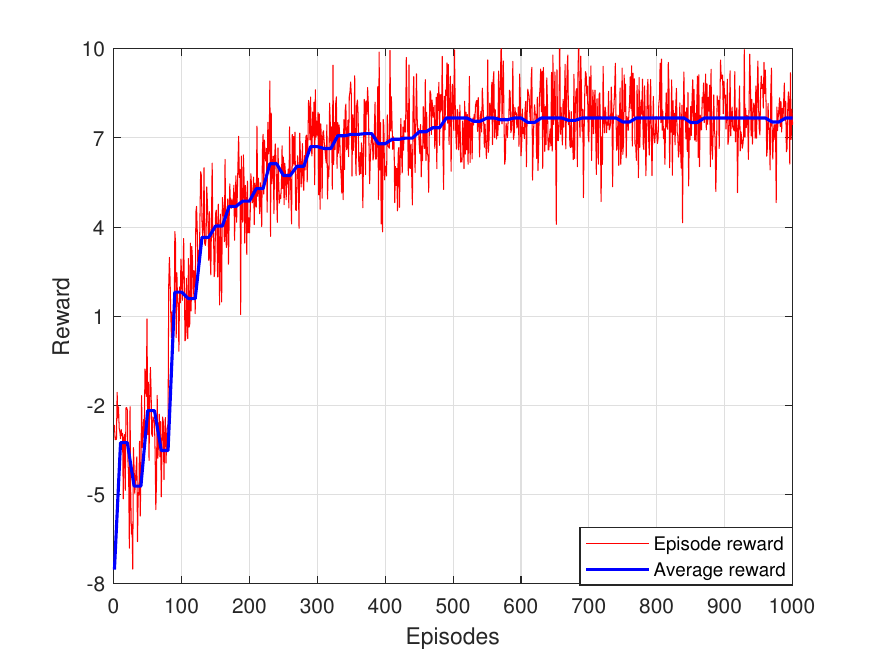}
		\caption{\color{black}The learning curves for the hybrid DQN-DDPG DRL model versus the number of training episodes.}\label{figTraining}
	\end{center} 
\end{figure}

\subsection{Results}
We compare the performance of the following   approaches:
\begin{itemize}
\item \textit{Hybrid DQN-DDPG  DRL}  which represents the results of the proposed DRL algorithm described in \textbf{Algorithm~1} that requires  a value of the object-preference weight $\zeta$.
\item \textit{The  meta-DRL}  which represents the results of the proposed meta-DRL algorithm described in \textbf{Algorithm~2}, \textbf{without}  the fine-tuning step.
\item \textit{The   meta-DRL with fine-tuning}  which represents the results of the proposed meta-DRL algorithm described in Section \ref{FineT} with  the fine-tuning stage in \textbf{Algorithm~3}.
\item \textit{The random solution} in which the decoding order and power allocation decisions are randomly selected.
\item {\textit{The exhaustive search solution} in which the power allocation decision is discretized into $ 10 $ levels and all the combinations of the decoding order and the discretized power allocation decisions are examined.}
\end{itemize}

 {
Figure \ref{figResZ1} illustrates the objective function versus the objective-preference
weight $\zeta$ for the proposed framework obtained using the
four solution approaches.  It is seen that   the    hybrid DQN-DDPG 
approach achieves near-optimum performance in comparison with the  exhaustive search solution     and the random approach provides the worst
performance. The performance of the meta-based DRL is better than that of the random solution and the fine-tuning stage improves the performance of the meta-based model and provides close to optimum solutions. It is worth mentioning that  a hybrid DQN-DDPG model is trained for each value of $\zeta$ in Fig. \ref{figResZ1}. 
\begin{figure}[ht]
	\begin{center}
		\includegraphics[width=1\linewidth]{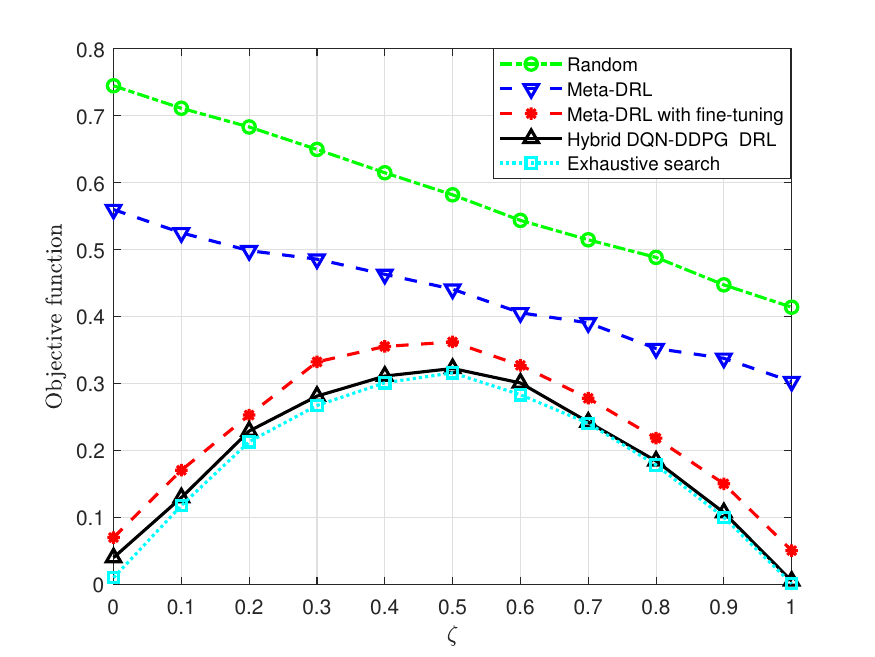}
		\caption{Objective function in \eqref{Obj}    versus  the relative weight $\zeta$ with  $V=10$ vehicles,   $F=4$   processes and $\abs{\mathcal{R}_i}=2$.}\label{figResZ1}
	\end{center} 
\end{figure}

To get more insight into this result, Fig. \ref{figResZ2} illustrates the corresponding  average
AoI and power consumption versus $\zeta$.  It is noticed that the
proposed framework provides a good trade-off between AoI
and power expenditure as for low values of $ \zeta $ it minimizes the
power expenditure and as $ \zeta $ increases it minimizes the AoI. The   trade-off   provided by the hybrid DQN-DDPG is close to that of the exhaustive search and the meta-DRL with fine-tuning provides a remarkably good performance.}  
That is not the case for the random solution, in which both
the AoI and power expenditure are not function of the relative
weight. It is worth noting the meta-DRL solution also not a function of $\zeta$ and its performance is   better than the random solution, as it is trained to minimize the objective function for   randomly sampled objective-preference weights.  
\begin{figure}[!ht]
	\begin{center}
			\includegraphics[width=1\linewidth]{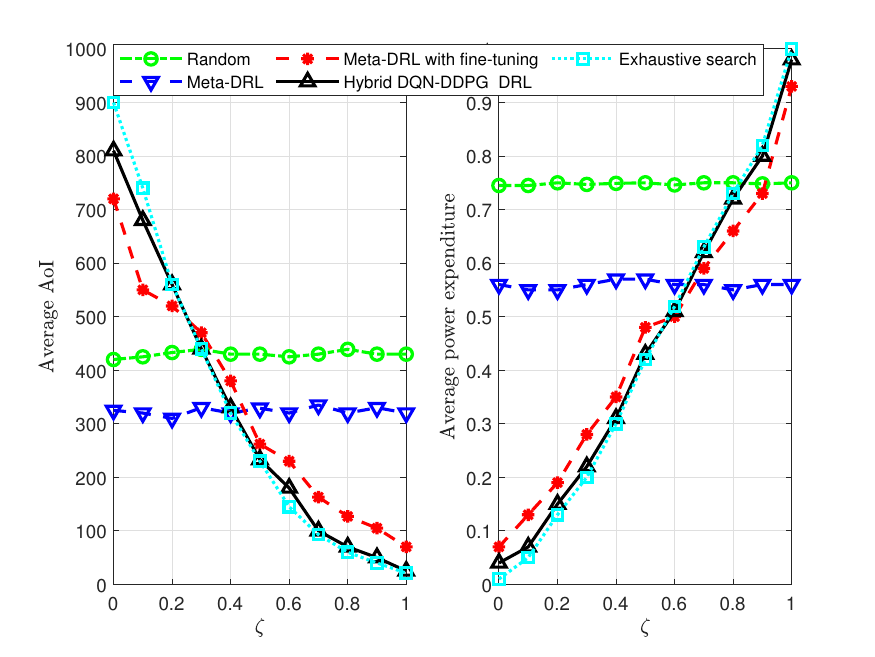}
			\caption{Average   AoI  and  power expenditure     versus  the relative weight $\zeta$ with  $V=10$ vehicles, processes $F=4$   processes and $\abs{\mathcal{R}_i}=2$.}\label{figResZ2}
		\end{center}  
\end{figure}

\begin{figure}[!ht]
	\begin{center}
		\includegraphics[width=1\linewidth]{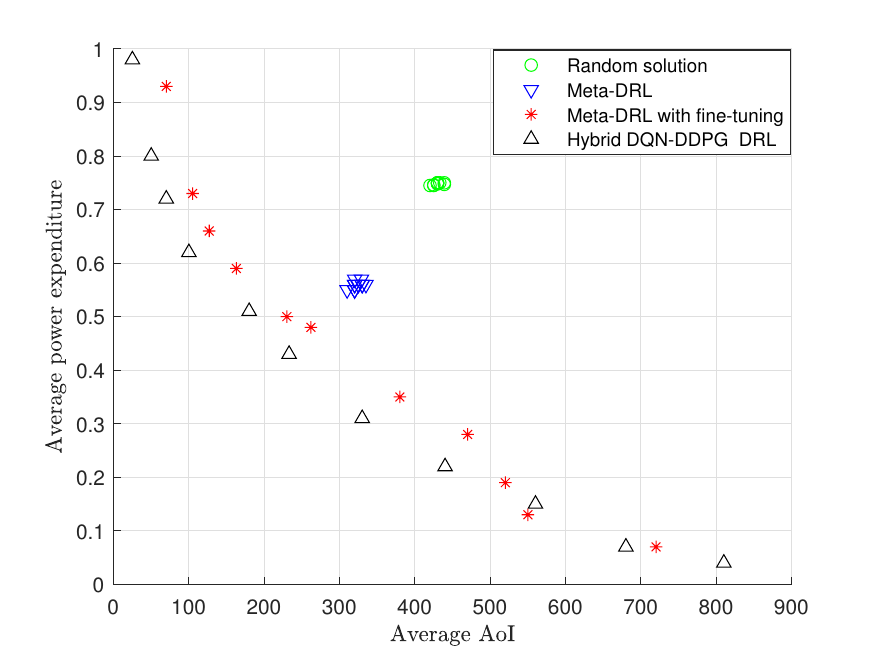}
		\caption{Pareto fronts with $V=10$ vehicles, $F=4$   processes,   $\abs{\mathcal{R}_i}=2$, and $\zeta=0:0.1:1$.}\label{figResPF}
	\end{center}  
\end{figure}
\begin{figure}[ht]
	\begin{center}
		\includegraphics[width=1\linewidth]{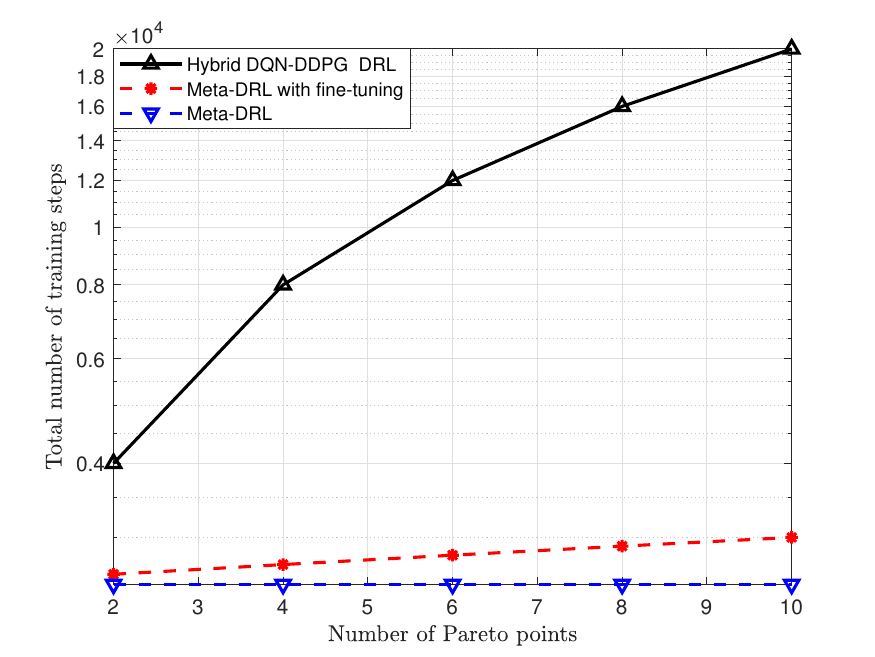}
		\caption{Total training steps to obtain the Pareto front.}\label{figResPF1}
	\end{center}  
\end{figure}

Figure \ref{figResPF}  illustrates the  Pareto fronts obtained by the four solution approaches. It can be noticed that the hybrid DQN-DDPG and the meta-DRL with fine-tuning approaches provide non-dominated solutions constitute   diverse and evenly spread of Pareto fronts, which gives the decision-maker a set of satisfactory trade-off solutions.  That is not the case for the random and meta-DRL approaches where the Pareto points are concentrated in close-proximity apart from the   lower left part of the
figure.  To quantify the quality of the
Pareto fronts in Fig. \ref{figResPF},   
the hypervolume  indicator (Lebesgue measure) of each Pareto front is calculated using Monte-Carlo approximation.
The hypervolume   indicator maps the set of points  in a Pareto front to  a single real value   measure that represents   the region
dominated by that Pareto front and  bounded above (for minimization problems) by a given reference point \cite{guerreiro2020hypervolume}. The higher the value of the hypervolume  indicator the better the  Pareto front to give the decision-maker satisfactory trade-off solutions. Assuming  the upper bound  of the two objectives  as a reference point i.e.,  (average AoI = $ \bar{\Delta}^{\mbox{\scriptsize max}}$, average power = $ P^{\mbox{\scriptsize max}} $), the hypervolume  indicators of the   random, meta-DRL, meta-DRL with fine-tuning, and
hybrid DQN-DDPG approaches in Fig. \ref{figResPF} equal \bm{$150$}, \bm{$315$}, \bm{$660$}, and \bm{$710$}, respectively.
It is worth mentioning that the desirable  performance of the  hybrid DQN-DDPG approach comes with a cost of training a model to obtain each Pareto point. However, the  meta-DRL with fine-tuning approach trains only one model and to obtain    a Pareto point it runs a few fine-tuning steps.

To illustrate the required training time to obtain the Pareto front, Fig. \ref{figResPF1} illustrates the total training time  versus the number of Pareto points. It is evident that the total training time  of the meta-based solutions is {significantly} less than that of the hybrid DQN-DDPG approach, {particularly for large number of Pareto points.} 
{Nonetheless, the  meta-DRL with fine-tuning approach achieves a Pareto front close to that of the hybrid DQN-DDPG approach, albeit having much less total training time.}

\begin{figure}[ht]
	\begin{center}
		\includegraphics[width=1\linewidth]{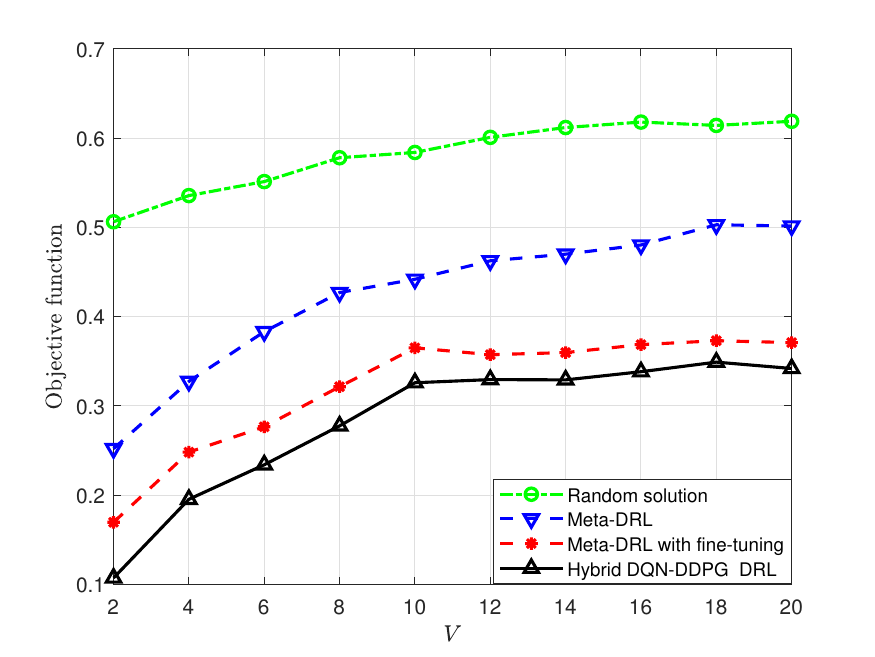}
		\caption{Objective function in \eqref{Obj}       versus  the number of vehicles with $V=10$ vehicles,   $F=4$   processes,   $\abs{\mathcal{R}_i}=2$, and $\zeta=0.5$ for the hybrid DQN-DDPG and meta-DRL with fine-tuning.}\label{figResV1}
	\end{center} 
\end{figure}
\begin{figure}[ht]
	\begin{center}
		\includegraphics[width=1\linewidth]{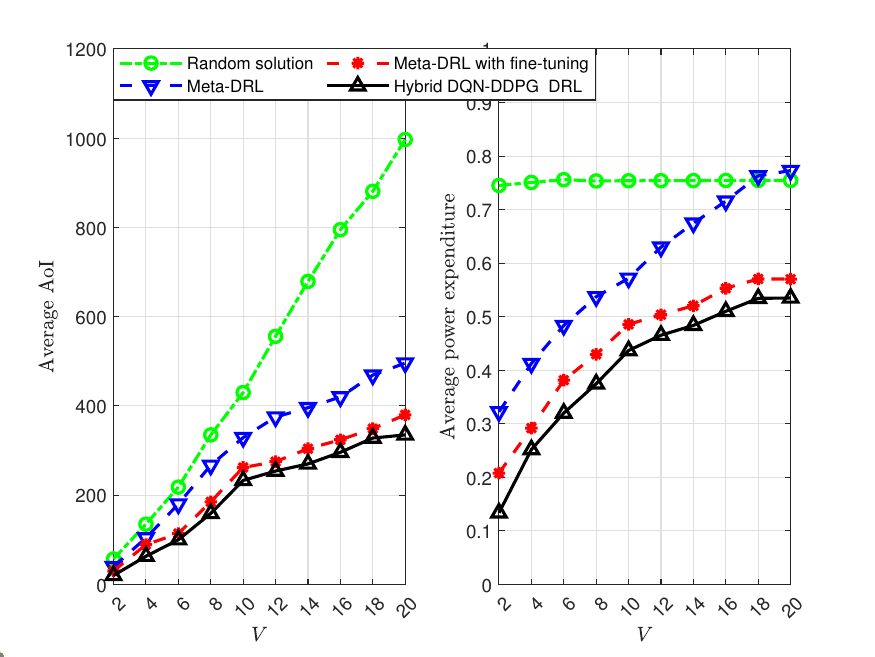}
		\caption{Average AoI and   power expenditure      versus  the number of vehicles with $V=10$ vehicles,   $F=4$   processes and $\abs{\mathcal{R}_i}=2$, and $\zeta=0.5$ for the hybrid DQN-DDPG and meta-DRL with fine-tuning.}\label{figResV2}
	\end{center} 
\end{figure}

\begin{figure}[!ht]
	\begin{center}
			\includegraphics[width=1\linewidth]{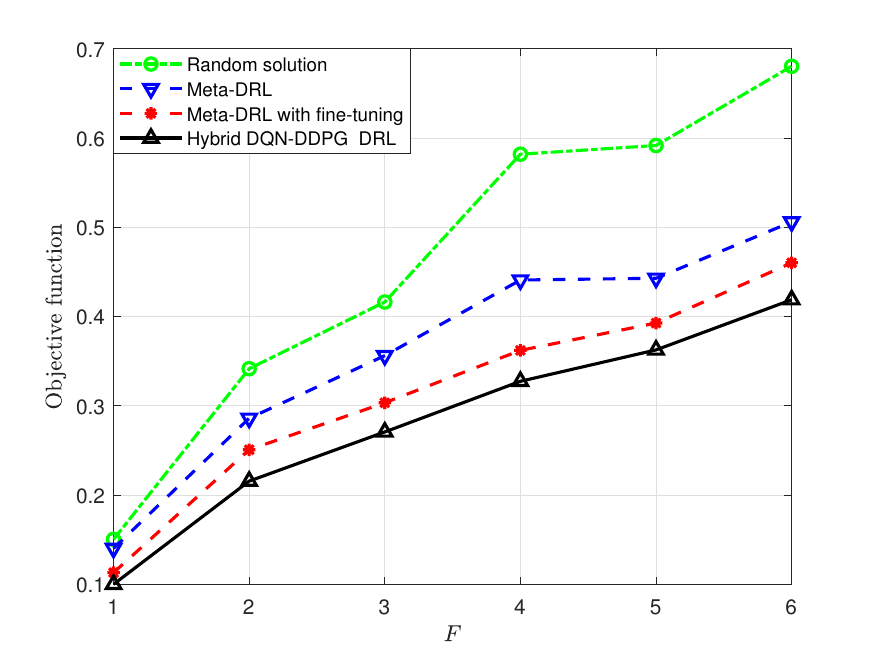}
			\caption{Objective function in \eqref{Obj}    versus the total number of processes $F$  with number of process of interest per vehicle $\abs{\mathcal{R}_i} =\ceil{\frac{F}{2}}$, $V=10$ vehicles,   and   $\zeta=0.5$ for the hybrid DQN-DDPG and meta-DRL with fine-tuning.}\label{figF1}
		\end{center}
\end{figure}  
\begin{figure}[!ht]
	\begin{center}
		\includegraphics[width=1\linewidth]{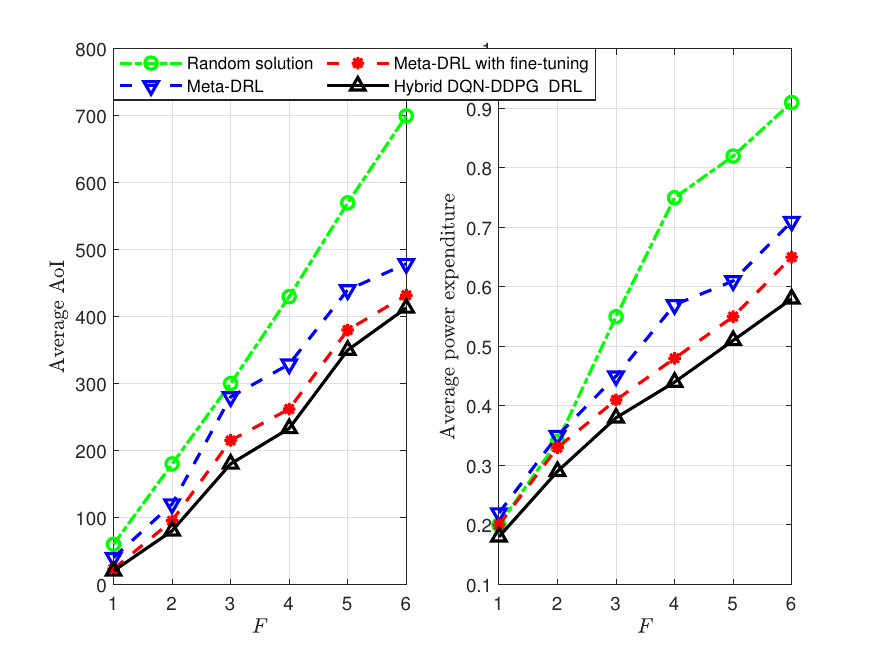}
		\caption{Average AoI and  power expenditure       versus the total number of processes $F$  with number of process of interest per vehicle $\abs{\mathcal{R}_i} =\ceil{\frac{F}{2}}$, $V=10$ vehicles,    and $\zeta=0.5$ for the hybrid DQN-DDPG and meta-DRL with fine-tuning.}\label{figF2}
	\end{center}
\end{figure} 

To illustrate the effectiveness of the proposed framework and solution approaches versus the number of vehicles $ V $, Fig. \ref{figResV1} and Fig. \ref{figResV2} illustrate the  objective function in \eqref{Obj} and the corresponding objectives over a  range of the number of vehicles, respectively. It can be seen that the proposed framework with the hybrid DQN-DDPG  DRL minimizes both the AoI and power consumption, while the AoI in the random solution increases rapidly as the number of vehicles increases. It can also noticed that the meta-DRL solution minimizes the  AoI and power consumption even though it was trained using randomly selected objective-preference
weights. The adaptation capability of the the meta-DRL model can be inferred as well as the   performance of the  meta-DRL with fine-tuning is close to that of the  conventional DRL.

Figures \ref{figF1} and \ref{figF2}
  illustrate     the  objective function in \eqref{Obj} and the corresponding average AoI and power consumption over a  range of the number of physical process, respectively. 
  It is worth mentioning that in these figures  as the number of physical process increases the number of required process per vehicle is also increases (i.e., $\abs{\mathcal{R}_i} =\ceil{\frac{F}{2}}$).  
 It can be noticed that both the AoI and power consumption increase as the number of physical process increases, and the meta-DRL with fine-tuning solution provides better  performance in comparison with the meta-DRL and random solutions.

{\color{black}
Figures \ref{figR1} and \ref{figR2}  show the performance of the solution approaches for a number of physical process   over a  range of the number of required process per vehicle.  It can be noticed that both the AoI and power consumption increase as the number of required process per vehicle increases, and the meta-DRL with fine-tuning solution provides better  performance in comparison with the meta-DRL and random solutions. 
}

\begin{figure}[!ht]
	\begin{center}
		\includegraphics[width=1\linewidth]{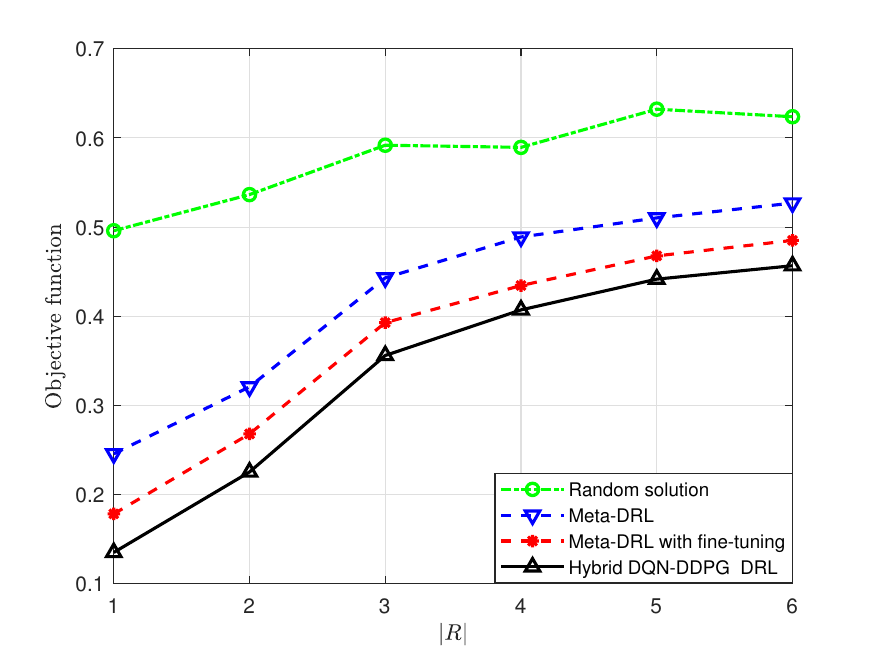}
		\caption{\color{black}Objective function in \eqref{Obj}     versus number of process of interest per vehicle $\abs{\mathcal{R}_i}$ with   $F=6$   processes, $V=10$ vehicles,  and   $\zeta=0.5$ for the hybrid DQN-DDPG and meta-DRL with fine-tuning.}\label{figR1}
	\end{center}
\end{figure}

\begin{figure}[!ht]
	\begin{center}
		\includegraphics[width=1\linewidth]{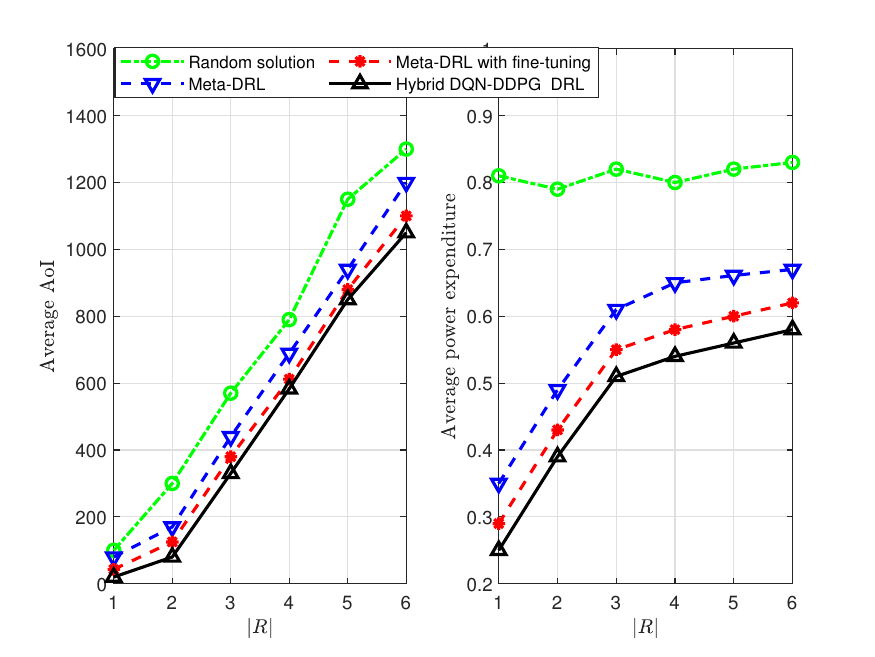}
		\caption{\color{black}Average  AoI  power expenditure    versus number of process of interest per vehicle $\abs{\mathcal{R}_i}$ with  $F=6$   processes, $V=10$ vehicles,  and $\zeta=0.5$ for the hybrid DQN-DDPG and meta-DRL with fine-tuning.}\label{figR2}
	\end{center}
\end{figure}

{\color{black}
To study the effect of the fine-tuning on the performance of the meta-based DRL solution, Fig. \ref{figTune} illustrates the performance of the  hybrid DQN-DDPG model, meta-based DRL, and meta-based DRL with fine-tuning solutions versus  the number of fine-tuning steps. The hybrid DQN-DDPG model is trained with the objective-preference weight of $ \zeta=0.5 $ while the meta-based DRL model is trained using   randomly selected values of $ \zeta $. It can be seen that as the number of the fine-tuning steps increases the performance of meta-DRL solution with fine-tuning improves.

\begin{figure}[ht]
	\begin{center}
		\includegraphics[width=1\linewidth]{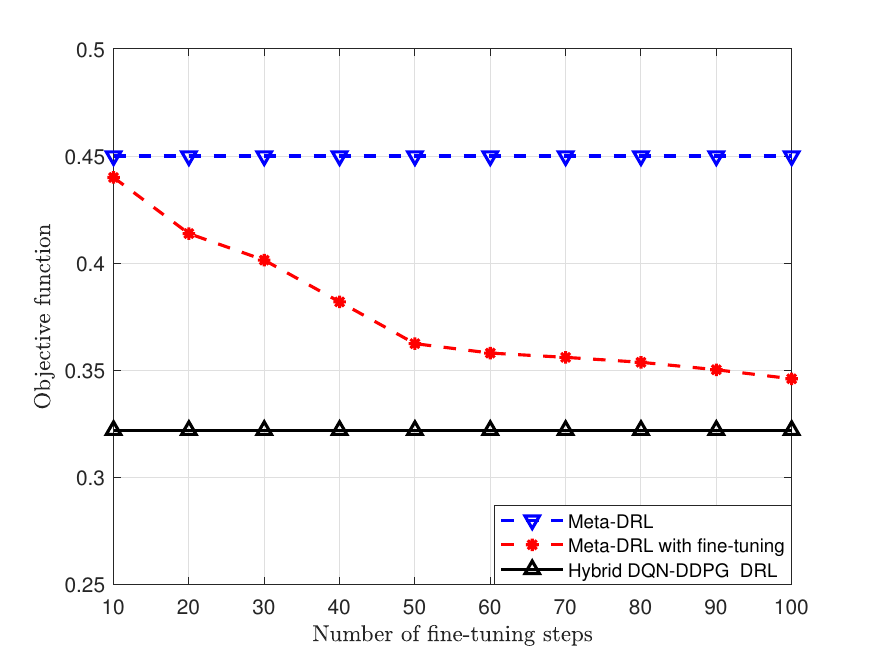}
		\caption{Objective function in \eqref{Obj}       versus  the number of fine-tuning steps with   $V=10$ vehicles, $F=4$   processes, $\abs{\mathcal{R}_i}=2$, and $\zeta=0.5$ for the hybrid DQN-DDPG and meta-DRL with fine-tuning.}\label{figTune}
	\end{center}   
\end{figure}

{
To  study the effect of the vehicles' mobility and the convergence  of the  meta-based DRL algorithm, Fig. \ref{figTuneS} illustrates the performance when
the DRL based model is trained with vehicle speed $ c_i \sim U(10, 15)  $ m/s  
and deployed on networks with three different vehicle speed values $20, 25$, and $30$ m/s. It can be noticed that the algorithm provides a good convergence performance and can adapt to the change in the network mobility after a few fine-tuning steps. It can be also inferred that the power required to deliver timely updates increases with the vehicles' speed, which consequently leads to a higher value of the objective function.
	\begin{figure}[!ht]
		\begin{center}
			\includegraphics[width=1\linewidth]{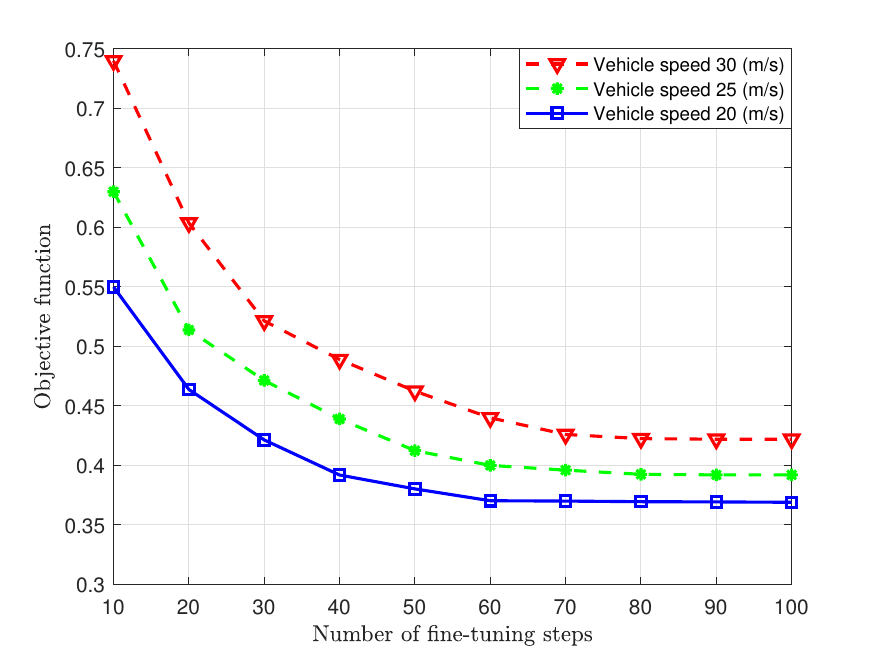}
			\caption{Objective function in (16)      versus  the number of fine-tuning steps    for vehicles' speeds $20, 25$, and $30$ m/s with $\zeta=0.5$,  $V=10$ vehicles, $F=4$   processes, and $\abs{\mathcal{R}_i}=2$.}\label{figTuneS}
		\end{center} 
	\end{figure}
}

\begin{figure}[!ht]
	\begin{center}
		\includegraphics[width=1\linewidth]{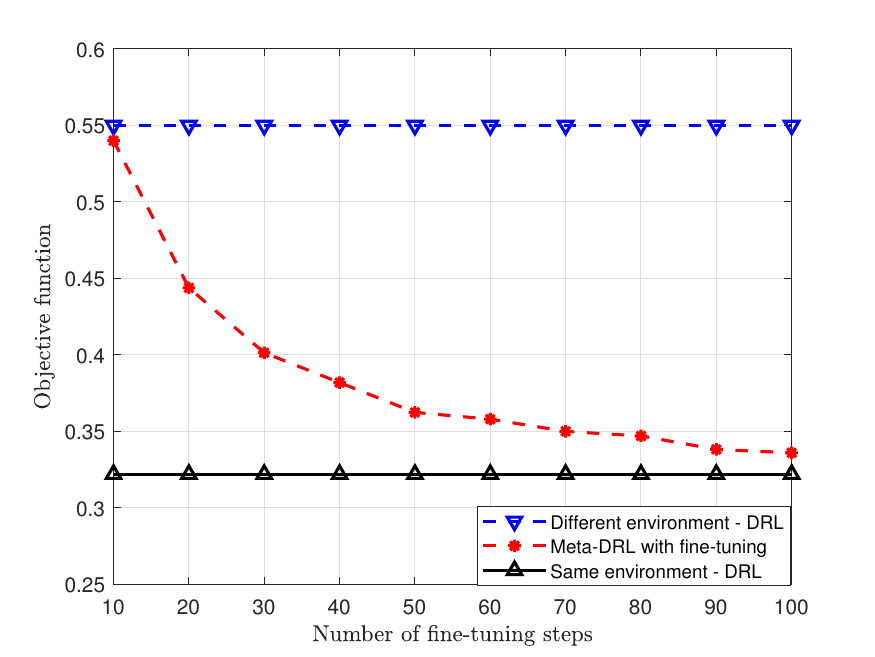}
		\caption{Objective function in \eqref{Obj}       versus  the number of fine-tuning steps with $\zeta=0.5$, total number of processes $F=4$   processes and $\abs{\mathcal{R}_i}=2$ for two different environments (1)   two lanes with $ c_i \sim U(10, 15)  $ m/s and (2)   four lanes with $ c_i \sim U(25, 35)  $ m/s.}\label{figTuneE}
	\end{center} 
\end{figure}

{
To  study the generalization ability of the  meta-based DRL algorithm to   new physical environments, we considered two different vehicular network scenarios in Fig. \ref{figTuneE}, namely, urban environment which considers a two lanes road with velocity of the vehicle is  $ c_i \sim U(10, 15)  $ m/s and  highway environment which considers a  four lanes  road  with velocity of the vehicle is $ c_i \sim U(25, 35)  $ m/s.
Fig. \ref{figTuneE} illustrates the performance of the  DRL model for  three different cases (1) the same-environment solution which represents the case when the DRL is trained and deployed on the same environment (i.e., the urban environment); (2) the different-environment solution which represents the case when the DRL is trained on the urban environment and deployed on the highway environment; and (3) the meta-DRL with fine-tuning solution which represents the case when the DRL is trained on the urban environment  fine-tuned and deployed on the highway environment. It can be seen that the meta-DRL model  has a
good generalization ability and can fast adapt to new environments via a few fine-tuning steps.}
 }

\section{Conclusion}\label{Con}
This paper has proposed an SIC-enabled message superposition framework to disseminate timely status updates about a set of physical processes to a set of vehicles. A  multi-objective mixed integer optimization problem has been formulated to minimize both the average AoI   and power consumption objectives in vehicular networks. 	 A hybrid  DQN-DDPG DRL    model has been developed  to solve the optimization
problem. To address the tradeoff between the  two objectives, 
  a meta-based DRL algorithm is trained using a range of objective-preference weights and undergoes a few fine-tuning update steps  to estimate the Pareto points  of the problem. Simulation results have illustrated that the hybrid DQN-DDPG solution minimizes both the AoI and power consumption for the given  objective-preference weight. Moreover, the meta-model demonstrates a rapid and
excellent adaptability to estimate  a high quality Pareto frontier  for all problem instances, including those with either unseen objective-preference weights or the new vehicular environments. {Considering DRL-based solution to handle more than two conflicting metrics in vehicular networks is considered as a future work.}

\bibliographystyle{IEEEtran}
\bibliography{IEEEabrv,Refrences-library}

\begin{IEEEbiography}
	[{\includegraphics[width=1in,height=1.25in,clip,keepaspectratio]{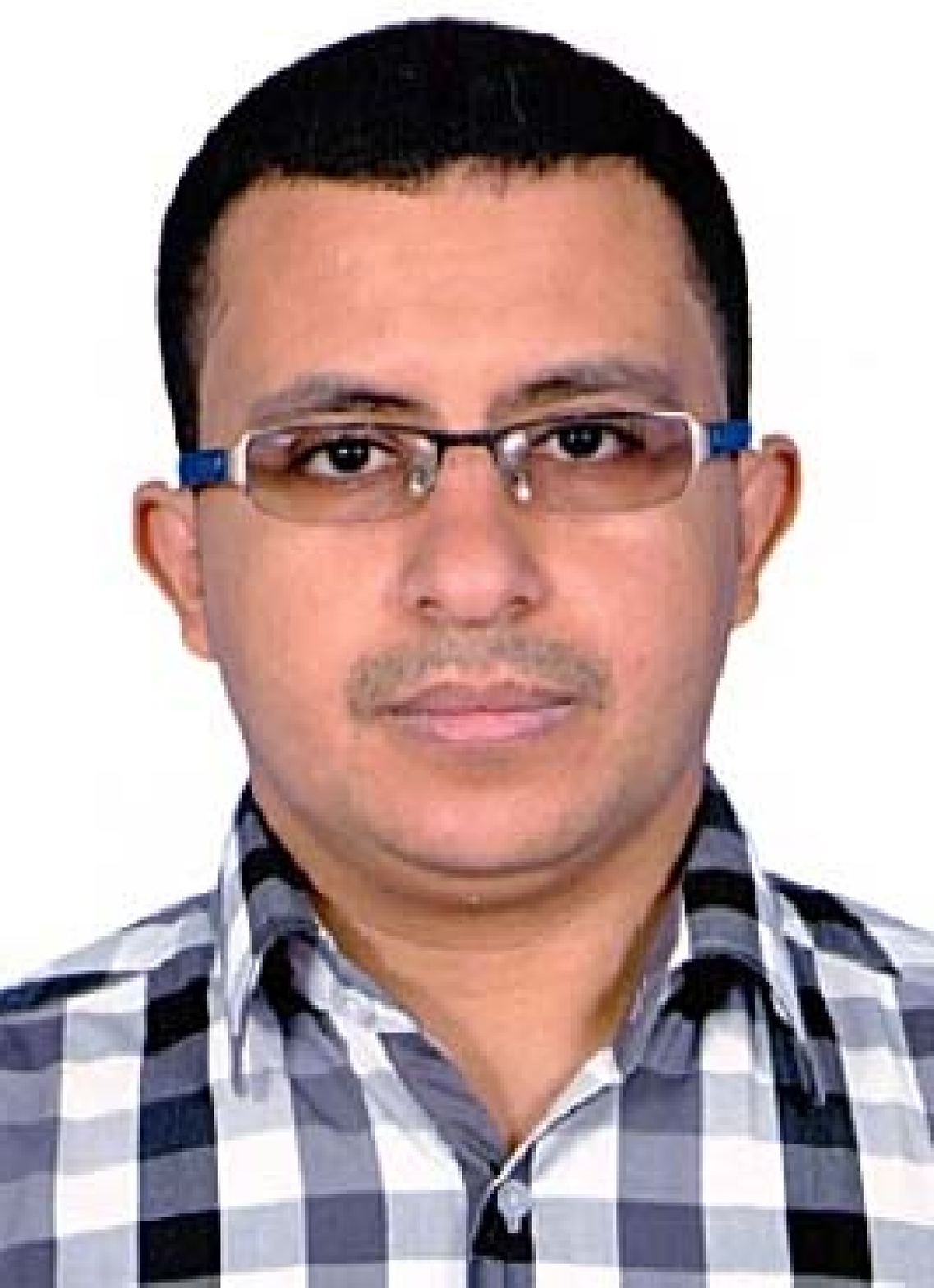}}]{Ahmed A. Al-habob}
	(S'15-M'22) received the BSc degree in telecommunications and computer engineering from Taiz University,\ Yemen,\ in 2009.\ He received the MSc degree in telecommunications from the Electrical Engineering Department, King Fahd University of Petroleum and Minerals (KFUPM), Saudi Arabia, in 2016. He received
	the Ph.D. degree in Electrical     Engineering from the 
	 Faculty of Engineering and Applied Science, Memorial University, St. John’s, NL, Canada, in 2022.\  He was a Postdoctoral Visitor   at Lassonde School of Engineering, York University, Toronto, ON, Canada.	  He is currently a Postdoctoral Fellow   at Memorial University of Newfoundland, NL, Canada.	 
	  His research interest includes wireless communications and networking.
\end{IEEEbiography}

\begin{IEEEbiography}
	[{\includegraphics[width=1in,height=1.25in,clip,keepaspectratio]{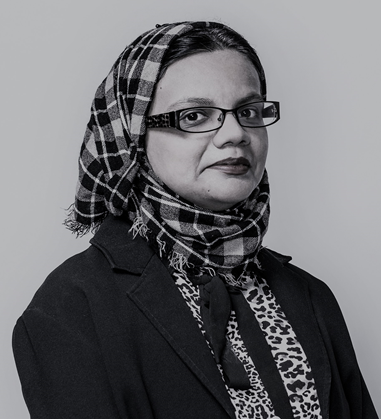}}]{Hina Tabassum}  (Senior Member, IEEE) received the Ph.D. degree from the King Abdullah University of Science and Technology (KAUST). She is currently an Associate Professor with the Lassonde School of Engineering, York University, Canada, where she joined as an Assistant Professor, in 2018. She is also appointed as the York Research Chair of 5G/6G-enabled mobility and sensing applications (2023 - 2028). She was a postdoctoral research associate at University of Manitoba, Canada.  She has published over 100 refereed papers in well-reputed IEEE journals, magazines, and conferences.  She received the Lassonde Innovation Early-Career Researcher Award in 2023 and the N2Women: Rising Stars in Computer Networking and Communications in 2022. She was listed in the Stanford’s list of the World’s Top $2\%$ Researchers in 2021, 2022, and 2023.  She is the Founding Chair of the Special Interest Group on THz communications in IEEE Communications Society (ComSoc)-Radio Communications Committee (RCC). She served as an Associate Editor for IEEE Communications Letters (2019–2023), IEEE Open Journal of the Communications Society (OJCOMS) (2019–2023), and IEEE Transactions on Green Communications and Networking (TGCN) (2020–2023). Currently, she is also serving as an Area Editor for IEEE OJCOMS and an Associate Editor for IEEE Transactions on Communications, IEEE Transactions on Wireless Communications, and IEEE Communications Surveys and Tutorials. She has been recognized as an Exemplary Editor by the IEEE Communications Letters (2020), IEEE OJCOMS (2023), and IEEE TGCN (2023).  Her research interests include multi-band optical, mm-wave, and THz networks and cutting-edge machine learning solutions for next generation wireless communication and sensing networks.
\end{IEEEbiography}

\begin{IEEEbiography}	[{\includegraphics[width=1in,height=1.25in,clip,keepaspectratio]{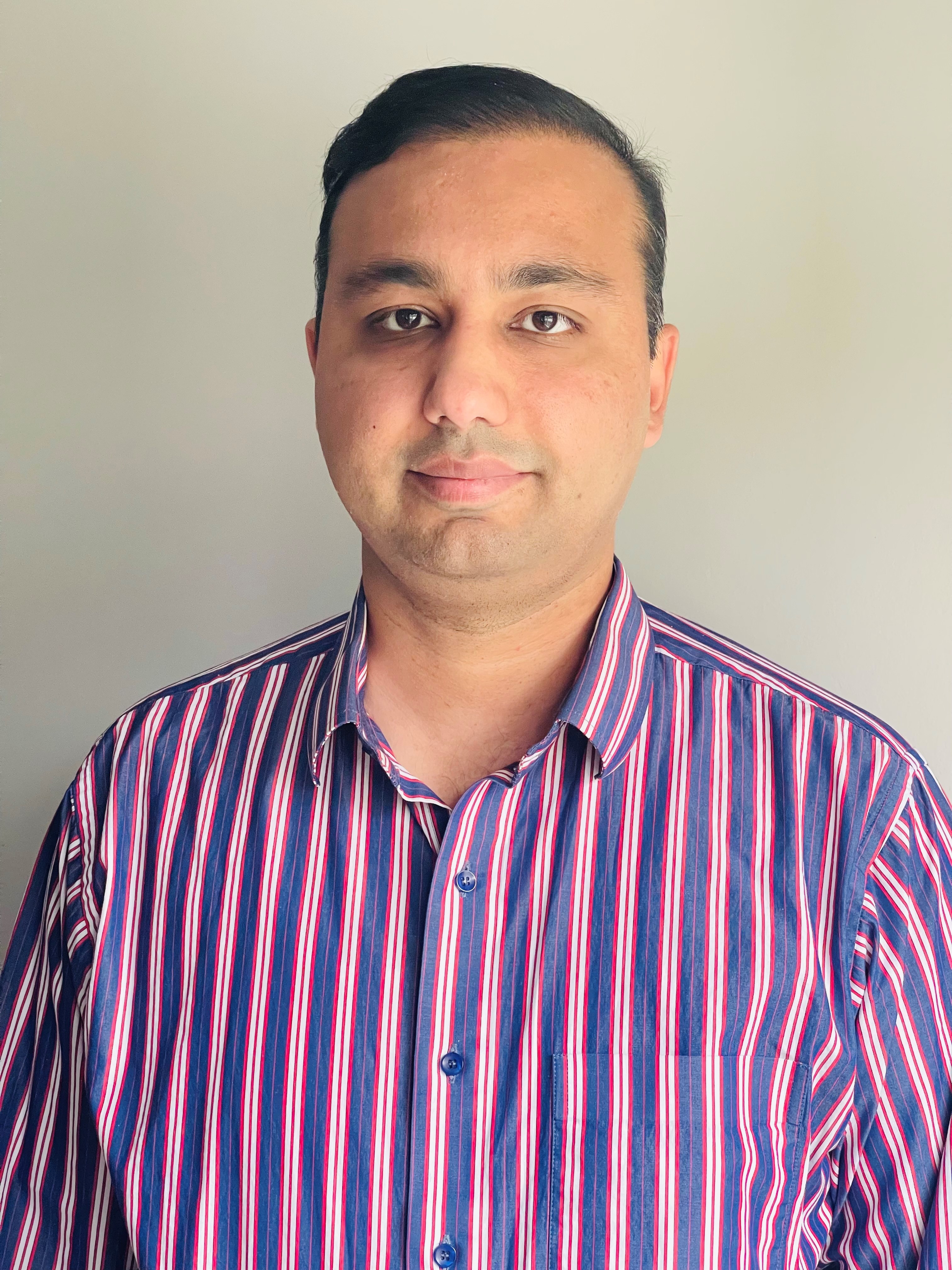}}]{Omer Waqar}
	received the B.Sc. degree in electrical engineering from the University of Engineering and Technology (UET), Lahore, Pakistan, in 2007 and the Ph.D. degree in electrical and electronic engineering from the University of Leeds, Leeds, U.K., in November 2011. From January 2012 to July 2013, he was a Research Fellow with the Center for Communications Systems Research and 5G Innovation Center (5GIC), University of Surrey, Guildford, U.K. He worked as an Assistant Professor in UET, Lahore, Pakistan from August 2013 to June 2018. He worked as a researcher in the department of Electrical and Computer Engineering, University of Toronto, Canada from July 2018 to June 2019. He worked as an Assistant Professor in the department of Engineering, Thompson Rivers University (TRU), British Columbia (BC), Canada from August 2019 to July 2023. Since August 2023, he has been working as an Assistant Professor in the School of Computing, University of the Fraser Valley, BC, Canada  and holds an adjunct faculty position at York university, Ontario, Canada. He has authored or co-authored 35+ peer-reviewed articles including top-tier journals such as IEEE Transactions on Vehicular Technology. He has secured over \$200K in research grants from the Tri-Council agency i.e., NSERC Discovery grant and NSERC Alliance grants. Currently, he is serving as an Associate Editor for the IEEE Open Journal of the Communications Society and IEEE Canadian Journal of Electrical and Computer Engineering. His current research interests include, intelligent reflecting surface aided communication systems, Deep-Learning for next generation communication networks, wireless sensing and resource allocation of wireless networks for several distributed machine learning paradigms.
\end{IEEEbiography}

\end{document}